%% file: main.tex
\documentclass[10pt,twocolumn,letterpaper]{article}

\usepackage{cvpr}              
\usepackage[accsupp]{axessibility}

\input{preamble}

\definecolor{cvprblue}{rgb}{0.21,0.49,0.74}
\usepackage{dsfont}
\usepackage{adjustbox}
\usepackage[pagebackref,breaklinks,colorlinks,citecolor=cvprblue]{hyperref}

\newcommand{\squeezeup}{\vspace{-2mm}}
\newcommand{\halfsqueezeup}{\vspace{-1mm}}

\def\paperID{15} 
\def\confName{CVPR}
\def\confYear{2024}

\title{PARASOL: Parametric Style Control for Diffusion Image Synthesis}

\author{Gemma Canet Tarrés$^1$, Dan Ruta$^1$, Tu Bui$^1$, John Collomosse$^{1, 2}$\\
$^1$University of Surrey, $^2$Adobe Research\\
{\tt\small \{g.canettarres, d.ruta, t.v.bui, j.collomosse\}@surrey.ac.uk}
}

\begin{document}
\twocolumn[{%
\renewcommand\twocolumn[1][]{#1}%
\maketitle
\vspace{-25pt}
\begin{center}
    \centering
    \includegraphics[width=\linewidth]{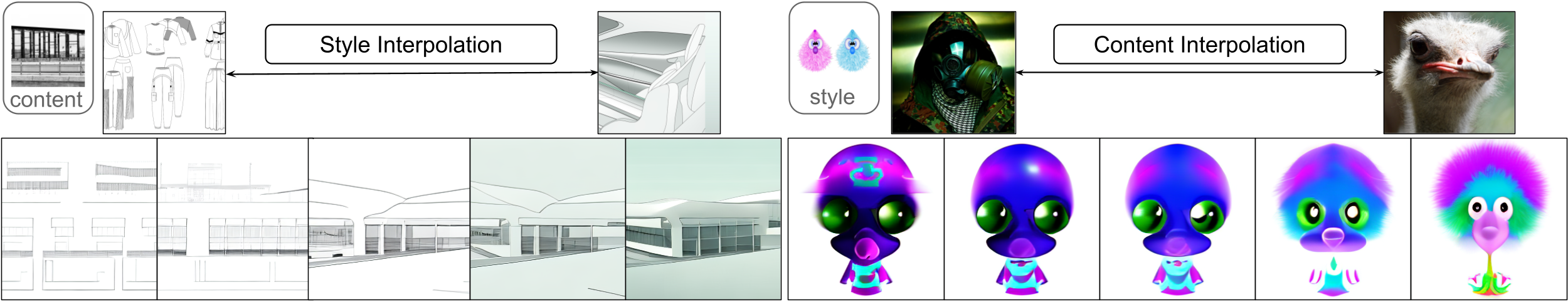}
Figure 1. Images generated using PARASOL. Leveraging parametric style control and multi-modal synthesis, we demonstrate the ability of our model to synthesize creative images by interpolating different styles (left) or contents (right).
    \label{fig:teaser}

\end{center}%

 \vspace{10pt}
 
}]
\setcounter{figure}{1}

\input{sec/0_abstract} 
\input{sec/1_intro}
\input{sec/2_method}
\input{sec/3_experiments}
\input{sec/4_conclusions}
{
    \small
    \bibliographystyle{ieeenat_fullname}
    \bibliography{main}
}

\input{sec/X_suppl}

\end{document}

%% file: preamble.tex
\usepackage[dvipsnames]{xcolor}


%% file: sec/0_abstract.tex
\begin{abstract}
We propose PARASOL, a multi-modal synthesis model that enables disentangled, parametric control of the visual style of the image by jointly conditioning synthesis on both content and a fine-grained visual style embedding. We train a latent diffusion model (LDM) using specific losses for each modality and adapt the classifer-free guidance for encouraging disentangled control over independent content and style modalities at inference time. We leverage auxiliary semantic and style-based search to create training triplets for supervision of the LDM, ensuring complementarity of content and style cues.  PARASOL shows promise for enabling nuanced control over visual style in diffusion models for image creation and stylization, as well as generative search where text-based search results may be adapted to more closely match user intent by interpolating both content and style descriptors.
\end{abstract}

%% file: sec/1_intro.tex
\section{Introduction}
\label{sec:intro}
\begin{figure}[t!]
    \includegraphics[width=\linewidth,trim=0cm 0cm 0cm 0cm,clip]{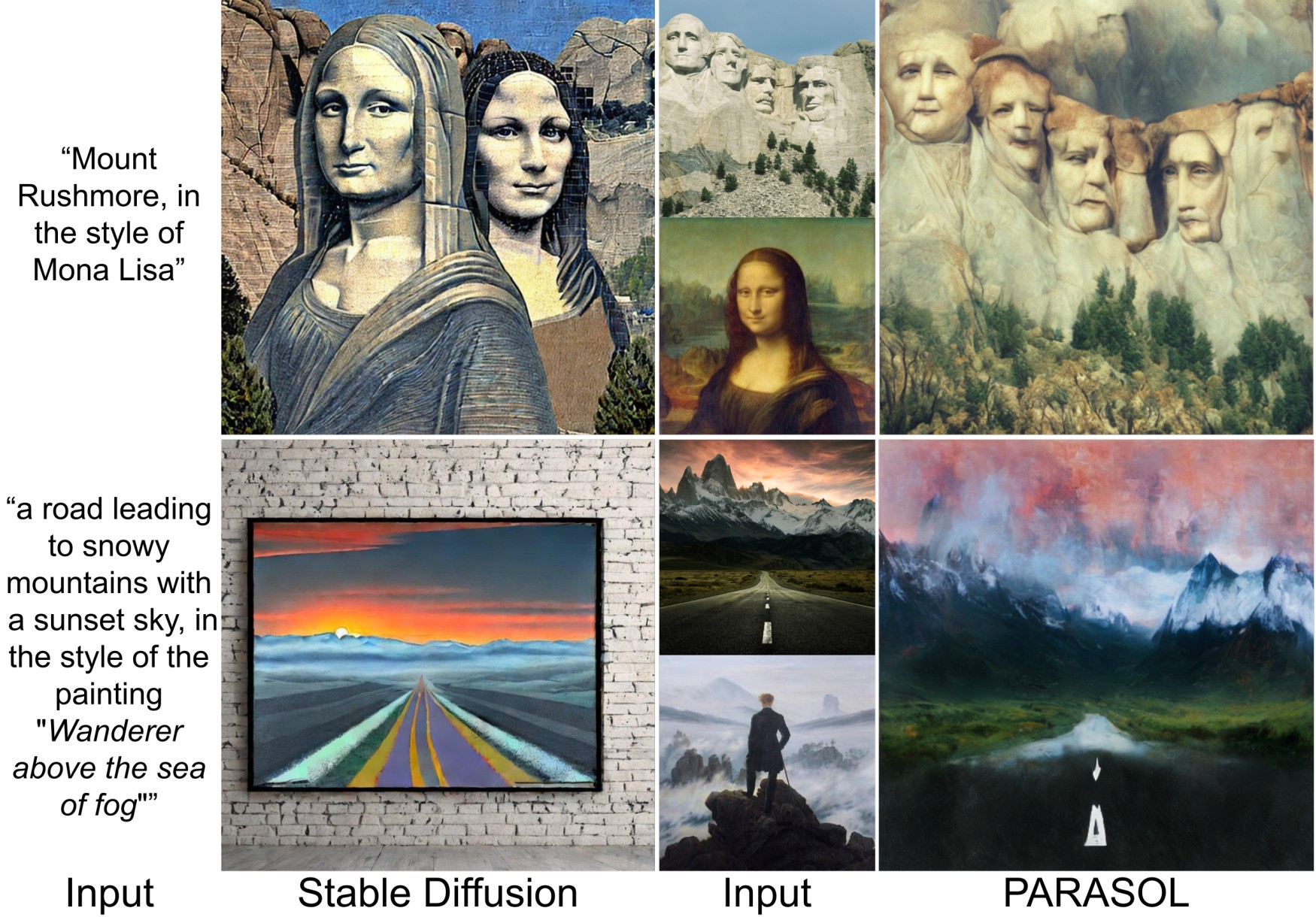}
    \caption{Comparison to Stable Diffusion \cite{ldm}. Stable Diffusion encounters difficulty in disentangling content and style as well as transferring the particular requested style, while PARASOL adeptly combines fine-grained details of both into its output.}
    \label{fig:famoussd}
    \vspace{-6mm}
\end{figure}

Deep generative models have immense potential for creative expression, yet their controllability remains limited. 
While diffusion models excel in synthesizing high-quality and diverse outputs, their fine-grained attribute control, especially in visual style, is often limited by coarse-grained inputs such as textual descriptions \cite{ldm}, structural visual cues \cite{controlnet} or style transfer \cite{cast,contraAST}. As shown in Fig. \ref{fig:famoussd}, these inputs present significant limitations: (i) they restrict the nuances that can be inherited from style inputs, (ii) without specifically disentangling both attributes, they hinder the model's ability to distinguish between content and style information. In contrast, visual search models often use \textit{parametric style embeddings} to achieve this more nuanced control. Leveraging such embeddings for guiding image synthesis, we propose Parametric Style Control (PARASOL) to bridge this gap. PARASOL is a novel synthesis model that enables disentangled parametric control over the fine-grained visual style and content of an image, conditioning synthesis on both a semantic cue and a fine-grained visual style embedding \cite{aladin}. We show how the use of parametric style embeddings also enable various applications, including (i) interpolation of multiple contents and/or styles (Fig. 1), (ii) refining generative search.
Additionally, for enhanced user control, we introduce test-time features in our pipeline that enable more control over the influence of each attribute on the output. Our approach holds relevance in real-world contexts such as fashion design, architectural rendering, and personalized content creation, where precise control over image style and content is essential for creative expression and practical utility. 
Thus, our technical contributions are:

{\textbf{Fine-grained style-conditioned diffusion.}  We synthesize images using a latent diffusion model (LDM) conditioned on multi-modal input describing independent content and style descriptors. A joint loss is introduced for encouraging disentangled control between the two modalities of interdependent nature. At inference time we invert the content image back to its noised latent, and re-run the denoising process incorporating content and style conditioning as well as modality-specific classifier-free guidance, enabling fine-grained control over the influence of each  modality.}

{\textbf{Cross-modal disentangled training.}  We use auxiliary semantic and style based search models to form triplets (content input, style input, image output) for supervision of the LDM training, ensuring complementarity of content and style cues to encourage disentangled control at inference.} 

{\textbf{Extensive evaluation} We evaluate our model thoroughly by comparing to conditional generation models and style transfer methods and show that PARASOL outperforms the state-of-the-art in various metrics and user studies.}



\halfsqueezeup
\section{Related Work}


\textbf{Style transfer and representation.}  Neural style transfer (NST) has classically relied upon aligning statistical features \cite{adain,wct} extracted from pretrained models (e.g. VGG \cite{gatys}) in the output image with those of an exemplar style image. More recently, style feature representations are learned via self-attention \cite{avatarnet,sanet}. PAMA \cite{pama} further improve upon this through progressive multi-stage alignment of features, improving consistency of style across the stylized image. ContraAST \cite{contraAST} introduce contrastive losses and domain-level adversarial losses to improve similarity of stylized images to real style images. CAST \cite{cast} refine this by including ground truth style images into the contrastive objective. InST \cite{zhang2023inversion} leverage inversion in diffusion models for transferring a new style.
NST methods aim to alter texture while retaining exact content of an image, hindering creativity and controllability in image generation. In this work, we aim at bridging this gap.

\textbf{Cross-modal/Style search.}
Early cross-modal retrieval work focused on canonical correlation analysis \cite{cca1,cca2}.  Deep metric learning approaches typically explore dual encoders unified via recurrent or convolutional layers  combined with metric learning over joint embeddings.  Recently transformers are popular in such frameworks, including BERT \cite{bert} and derivatives for language and ViT \cite{vit} based image representations. Vision-language models (e.g. CLIP \cite{CLIP}) have been shown effective for search and conditional synthesis.  

In terms of style, earlier style transfer work \cite{old_style_rep} explore representation learning of artistic style in the context of fine art, learning 10 styles to generalize a style transfer model to more than a single style. \cite{Collomosse2017} leverage a labelled triplet loss to learn a metric style representation over a subset of the BAM dataset \cite{bam}, but are limited by the 7 available style labels in BAM. ALADIN \cite{aladin} first explored a fine-grained style representation learning through multi-layer AdaIN \cite{adain} feature extraction, trained over their newly introduced BAM-FG dataset. This was evolved in StyleBabel \cite{stylebabel} by replacing the architecture with a vision transformer \cite{vit}, which we use in our work.


\textbf{Multi-modal conditional image generation.} Due to the outstanding quality of recent image generation methods, conditional image generation is currently becoming a focus of attention in research. These aim to achieve a more controllable synthesis. Most conditional methods consider a single input modality ($\eg$ text  \cite{imagen,glide,ldm,dalle2,vqdiff}, bounding box layouts \cite{Sylvain2020,Zhao2018,Wei2019}, scene graphs \cite{sg1,Ashual2019}) and a few accept multi-modal conditions. VAEs \cite{VAE_real} have often been used for this task, since they allow learning a joint distribution over several modalities while enabling joint inference given a subset \cite{POE68_mvae, POE54_mmvae, POE28}.

GAN-based methods follow different conditioning strategies. TediGAN \cite{POE69_tedigan} relies on a pretrained unconditional generator, IC-GAN \cite{icgan} is based on a conditional GAN that leverages search for synthesizing new images. PoE-GAN \cite{huang2021poegan} uses a product of experts GAN to combine several modalities (sketch, semantic map, text) into an image. 

Make-A-Scene \cite{makeascene} and CoGS \cite{cogs} leverage the benefits of transformers and learned codebooks for incorporating multimodal conditioning. They both encode each modality independently into discretized embeddings that are modelled jointly by the transformer. 

More recently, several multi-modal diffusion-based image generation methods have been proposed. DiffuseIT \cite{diffuseit} and CDCD \cite{contrastivediff} combine the different modalities by tuning specific losses. Others make use of multimodal embeddings, such as CLIP \cite{CLIP} for combining text and image-based inputs \cite{drawyourart} or even retrieve auxiliary similar images to further condition the network \cite{rdm,knndiff,rdmeccv}. eDiffi \cite{ediffi} uses specific single-modality encoders and incorporates all embeddings through cross-attention at multiple resolutions. ControlNet \cite{controlnet} and MCM \cite{cusuh2023modulating} explore the integration of new input modalities as extra conditioning signals by leveraging the frozen Stable Diffusion model \cite{ldm}. In contrast, we train our latent diffusion with explicit governing loss for each input modality (i.e. content and style).


\squeezeup

%% file: sec/2_method.tex
\section{Methodology}
\label{sec:formatting}

We propose PARASOL; a method to creatively synthesize new images with disentangled parametric control over the fine-grained visual style and content. The main design choices making that possible are as follows:


\begin{itemize}
    \item By incorporating a \textbf{parametric style encoder} in our pipeline, our model is able to incorporate a more \textit{fine-grained style} to the generated image. 
    \item An \textbf{inverse diffusion step} is incorporated at sampling time for enabling control over the amount of \textit{content details and structure} to be preserved. 
    \item A specific \textbf{classifier-free guidance} format is introduced for each modality to \textit{independently influence} the output.
    \item The \textbf{metric properties} of both modality-specific encoders allow multiple styles and/or semantics to be \textit{combined and interpolated} for generating creative content.
\end{itemize}

\halfsqueezeup

\begin{figure*}[t]
    \centering
    \includegraphics[width=0.95\textwidth]{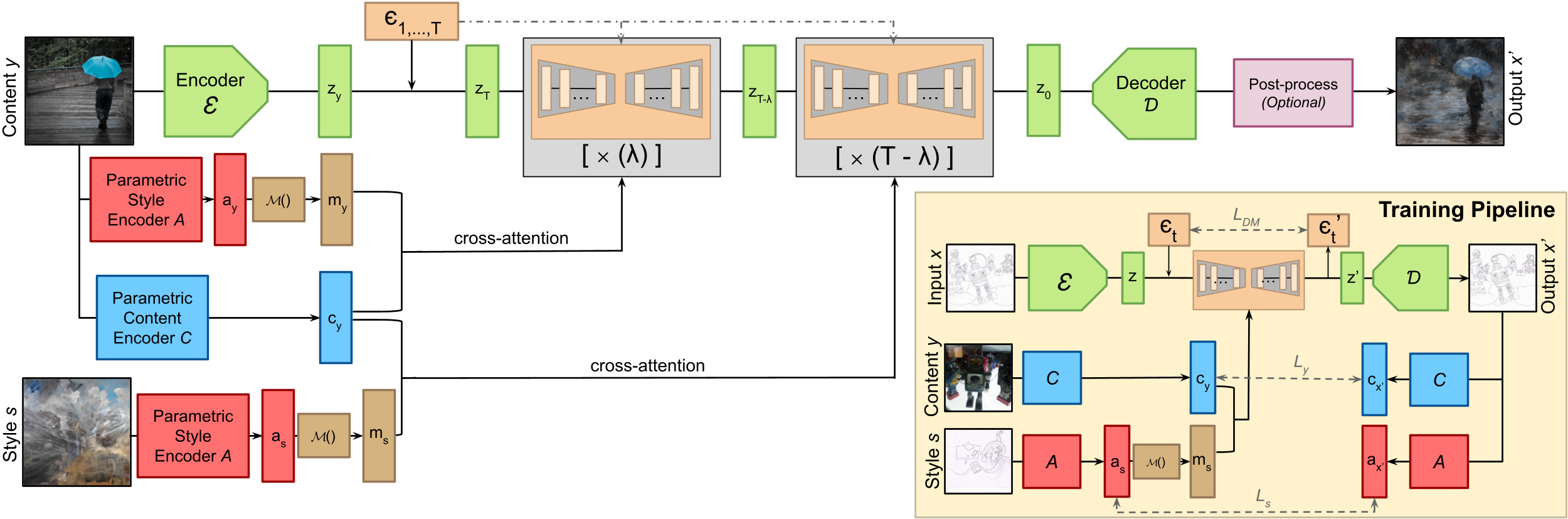}
    \squeezeup
    \caption{Illustration of the full PARASOL pipeline. It consists of six components: A parametric style encoder $A$ (red); A projector network $\mathcal{M}()$ (brown); A semantics encoder $C$ (blue); An Autoencoder $(\mathcal{E}, \mathcal{D})$ (green); A denoising U-Net (orange); An optional post-processing step (purple). At training time (bottom-right corner), two modality-specific losses ($L_s$ and $L_y$) are used to encourage disentanglement. They are combined with $L_{DM}$ and minimized in the training. At inference time (big pipeline), a parameter $\lambda \in [0, T]$ is introduced. After $\lambda$ denoising steps, the style condition is changed for transferring a new style.}
    \label{fig:pipeline}
    \squeezeup
    \squeezeup
\end{figure*}

The core of PARASOL is a pre-trained LDM \cite{ldm} fine-tuned for accommodating multimodal conditions. The pipeline (Fig. \ref{fig:pipeline}) consists of six components: 1) An Autoencoder ($\mathcal{E}$, $\mathcal{D}$) for encoding/decoding images into the diffusion latent space; 2) A U-Net based denoising network; 3) A parametric style encoder $A$ enabling fine-grained control over visual appearance; 4) A semantic encoder $C$ to express content control; 5) A projector network $\mathcal{M}()$ to bridge both modality embeddings into a same feature space; 6) An optional post-processing step for colour correction.


\subsection{Obtaining Training Supervision Data via Cross-Modal Search}
\label{seq:dataset}

A dataset of triplets ($x$, $y$, $s$) is required to train PARASOL. In each set, the output image $x$ is a stylistic image matching the artistic style of the style image $s$, and semantic content of the content image $y$. 

This dataset is built via cross-modal search for the input modalities: content and style.
Given an image $x$, its semantics descriptor $c_x = C(x)$ and style descriptor $a_x = A(x)$ are computed using parametric modality specific encoders ($C$ and $A$). Leveraging their parametric properties, the most similar images for each modality can be retrieved by finding the nearest neighbours in the respective feature spaces.

Certain restrictions are applied to ensure disentanglement between both input modalities. First, different data is indexed for each modality's search. A set of images with stylistic and aesthetic properties is defined as the ``Style Database" ($\mathcal{S}$). These are indexed using a parametric style encoder and used to find the style image $s$. In parallel, a set of photorealistic images with varied content is defined as the ``Semantics Database" ($\mathcal{C}$) and is indexed using a parametric semantics encoder for finding the content image $y$.

Using $a_x$ as query, the top $k$ most style similar images in $\mathcal{S}$ are retrieved as candidates for the style image $s$. Their similarity to $x$ in the semantics feature space is computed and they are discarded as candidates if this similarity is over a certain threshold. Finally, the style image $s$ is picked as the closest image in $\mathcal{S}$ fullfilling all restrictions. Using the semantics description $c_y$ as query, same procedure is conducted for finding the content image $y$ in $\mathcal{C}$. Visualization of such triplets can be found in SuppMat. 

\halfsqueezeup
\subsection{Encoding the Style and Semantics Inputs}

Given a triplet (content image $y$, style image $s$, output image $x$), the diffusion model is trained to reconstruct $x$ by conditioning on $y$ and $s$, after encoding the fine-grained style and content inputs through the respective encoders. Opting for $y$ over $x$ as the content image enables better style matching, acknowledging that perfect content alignment isn't always desirable (e.g., transferring a childish sketch style or a hand-drawn diagram requires changes in overall content shapes). Therefore, training the network using $y$, which has similar content structure and semantics to $x$, enables flexibility to accommodate a wider range of styles.

Arguably, the same encoder could be used on $s$ and $y$ to describe both style and content modalities, as in \cite{rdm}. However, the use of modality-specific encoders pre-trained on a task-specific curated data has been shown to be beneficial for conditioning the diffusion process \cite{imagen}. Furthermore, using a condition-specific encoder (i.e. style or semantics specific) strongly contributes to disentangling the content and style features. 
Nonetheless, using a different encoder for each modality poses a new challenge. When each modality is encoded into a distinct feature space, the LDM must be conditioned on both. Fine-tuning an LDM that was pre-trained on one conditional modality to understand and incorporate the information of two separate ones can be very data and compute demanding. Hence, we train an MLP-based \textit{projector network} $\mathcal{M}()$ to obtain a joint space for both descriptors. $\mathcal{M}()$ takes the style descriptor $a_s$ of $s$ as input and returns $m_s$, a new embedding that lies in the same feature space as $c_y$, the semantics descriptor of $y$. This allows for the LDM to be easily conditioned on both modalities while maintaining the disentanglement and details encoded in their respective embeddings.

\halfsqueezeup

\subsection{Incorporating Latent Diffusion Models}

PARASOL is built with a diffusion backbone. This LDM has two components: An Autoencoder (consisting of an Encoder $\mathcal{E}$ and a Decoder $\mathcal{D}$) and a U-Net denoising network. 
For each image $x$, the encoder $\mathcal{E}$ embeds it into $z = \mathcal{E}(x)$, while the decoder $\mathcal{D}$ can reconstruct $x' = \mathcal{D}(z)$ from $z$. 
The diffusion process takes place in the Autoencoder's latent space. This process  can be interpreted as a sequence of denoising autoencoders $\epsilon_\theta (z_t, t)$ that estimate $z_{t-1}$ from its noisier version $z_{t}$. 

\vspace{-5mm}
\paragraph{Conditioning through Cross-Attention} During our training, we freeze $(\mathcal{E}, \mathcal{D})$ and fine-tune the U-Net to incorporate multimodal conditions. In particular, the diffusion process needs to be conditioned on two independent signals: $m_s$ and $c_y$. We adapt the denoising autoencoder $\epsilon_\theta$ to condition on both signals as well as on the noisy sample $z_t$ and the timestamp $t$: $\epsilon_\theta (z_t, t, m_s, c_y)$. Inspired by \cite{ldm,rdm}, we incorporate these new signals into the U-Net backbone through the use of cross-attention by stacking the two signals $m_s$ and $c_y$ together and mapping them into every intermediate layer of the U-Net via cross-attention layers. As a result, at each timestep $t$, the output $z_{t-1}$ of the model $\epsilon_\theta$ is computed taking both conditions $m_s$ and $c_y$ into account.

\vspace{-5mm}
\paragraph{Conditioning using Classifier-free Guidance}
As presented in \cite{glide,dhariwal2021}, samples from conditional diffusion models can be improved by the use of a classifier(-free) guidance. The mean and variance of the diffusion model are then additively perturbed by the gradient of the log-probability of the conditioning modality leading to the generation of a particular image.

We extend the idea of classifier-free guidance for \textit{independently} accommodating multiple modalities. At training time, both input conditions are replaced by a null condition $\epsilon_\theta (z_t, t, \emptyset, \emptyset)$ with a fixed probability so that the network can learn how to produce unconditional outputs. Then, at sampling time, this output is guided towards $\epsilon_\theta (z_t, t, m_s, \emptyset)$ and $\epsilon_\theta (z_t, t, \emptyset, c_y)$ and away from $\epsilon_\theta (z_t, t, \emptyset, \emptyset)$ as:

\squeezeup

\begin{equation}
    \begin{aligned}
            \epsilon_\theta (z_t, t, m_s, c_y) = \epsilon_\theta (z_t, t, \emptyset, \emptyset) \\
            + g_s \big[ \epsilon_\theta (z_t, t, m_s, \emptyset) - \epsilon_\theta (z_t, t, \emptyset, \emptyset) \big] \\
            + g_y \big[ \epsilon_\theta (z_t, t, \emptyset, c_y) - \epsilon_\theta (z_t, t, \emptyset, \emptyset) \big]
    \end{aligned}
    \label{eq:gamma}
\end{equation}

The parameters $g_s$ and $g_y$ are introduced to determine how much weight the style or semantics inputs have in the image generation process. Thus, by tuning the ratio of the two parameters $g_s$ and $g_y$, the user can approximate the degree of influence the style and semantics inputs have over the image synthesis process. However, it should be noted that high values of either parameter will lead to higher quality and lower diversity.

\halfsqueezeup
\subsection{Training Pipeline}

At each training step, a random timestep $t \in [1,T]$ is selected. As shown in Fig. \ref{fig:pipeline} (bottom-right), each training image $x$ is encoded into $z$ using the pre-trained encoder $\mathcal{E}$ and noised with Gaussian noise $\epsilon_t$. Similar images in terms of style $s$ and content $y$ are encoded into $m_s$ and $c_y$. The embedding $z_t$ is then fed into the U-Net denoising autoencoder, which is also conditioned on $m_s$ and $c_y$ through cross-attention. We simultaneously train the U-Net and the projector network $\mathcal{M}()$ while freezing all other modules. 

\vspace{-4mm}
\paragraph{Training Objectives}
Training is done by minimizing a combination of 3 losses:

\textit{Diffusion Loss:} Through a reparametrization trick \cite{ddim,ldm}, minimizing the distance between the predicted noise $\epsilon_t ' := \epsilon_\theta (z_t, t)$ and the true noise $\epsilon_t$ is equivalent to minimizing the distance between $z_t$ and $z_{t-1}$. Thus, the loss to minimize is:

\squeezeup
\begin{equation}
    L_{DM} = \mathds{E}_{z_t, \epsilon_t \sim \mathcal{N} (0,1), t} \big[ \| \epsilon_t - \epsilon'_t \|^2 \big].
\label{eq:loss_diff}
\end{equation}
\halfsqueezeup


\textit{Modality-Specific Losses.} For encouraging the output image to have the same style as $s$ and the same semantics as $y$, the style and semantics of the reconstructed image $x'$ are encoded through the use of a style and a semantics encoder, respectively. The two losses are thus computed as: 
\begin{equation}
    L_s = \text{MSE}(a_s, a_{x'}), L_y = \text{MSE}(c_y, c_{x'}).
\end{equation}

\textit{Total Loss.} All three losses are combined in a weighted sum and simultaneously optimized:
\squeezeup
\begin{equation}
    L = L_{DM} + \omega_s \cdot L_s + \omega_y \cdot L_y,
    \label{eq:loss_total}
\end{equation}

with $\omega_s$ and $\omega_y$ being two weight parameters.

\subsection{Sampling Pipeline} \label{seq:sampling}

When sampling (Fig. \ref{fig:pipeline}), an inversion process \cite{wu2023uncovering} takes place to ensure the fine-grained content details in $y$ can be preserved in the final image. First, the semantics image $y$ is encoded via $\mathcal{E}$ and noised through a complete forward diffusion process. During this process, the noise $\epsilon_t$ introduced at each $t=1,...,T$ is being saved. These $\epsilon_t$ values are then used for sequentially denoising the image in the reverse diffusion process.
If the input conditions are unchanged throughout the whole denoising process, the image $y$ is faithfully reconstructed. For offering the possibility of transferring a new style while preserving all fine-grained content details in the image, $\lambda \in [1, T]$ is introduced. In the first $\lambda$ denoising steps, the U-Net is conditioned through cross-attention on the style and semantics descriptors of $y$, while in the last $T - \lambda$ steps, the style condition is switched to $m_s$, the encoded style from $s$ (\ie, for $T=50$, $\lambda=20$, the style and semantics of $y$ are used in $20$ steps and the target style in the remaining $30$). Therefore, setting $\lambda$ close to $T$ leads to more structurally similar images to $y$ while lower $\lambda$ values generate images stylistically closer to $s$. 



\vspace{-3mm}
\paragraph{Colour Distribution Post-Processing}
A challenge with perceptually matching the style of a reference image $s$ through diffusion, is mis-matched colour distribution, despite image fidelity to $y$. We offer the possibility of addressing this issue via additional post-processing steps inspired by ARF \cite{arf}. In this optional step, the generated image is modified to match the mean and covariance of the style image, shifting the colour distribution. 

\squeezeup

%% file: sec/3_experiments.tex
\vspace{-1mm}
\section{Experiments}
\halfsqueezeup

We discuss our experimental setup and evaluation metrics in Section \ref{seq:experimental_setup} and compare to baselines (Section \ref{seq:baselines}) and different ablations (Section \ref{seq:ablation}). We showcase user control of PARASOL in experiments in Section \ref{seq:controllability}, and its use for interpolation and as a generative search model for improving fine-grained control is investigated in Section \ref{seq:applications}. See SuppMat for more experiments and visualizations.

\halfsqueezeup
\vspace{-1mm}
\subsection{Experimental Setup} \label{seq:experimental_setup}

\textbf{Network and training parameters.} We use a pre-trained ALADIN \cite{aladin} as parametric style encoder and the pre-trained ``ViT-L/14" CLIP \cite{CLIP} for encoding the content input. The Autoencoder and U-Net are used from \cite{rdm}. The U-Net is fine-tuned using a multimodal loss (Eq. \ref{eq:loss_total}) with weights $\omega_s = 10^5$ and $\omega_y = 10^2$. At sampling time, unless stated otherwise, we use $\lambda = 20$, $g_s = 5.0$, and $g_y = 5.0$ and no post-processing in our experiments and figures. The training of PARASOL takes $\sim 10$ days on an 80GB A100 GPU, while the sampling of an image takes $\sim 5-90$s depending on $T$ and $\lambda$.

\textbf{Dataset.} The final model and all the ablations are trained using our own set of 500k triplets obtained as described in Section \ref{seq:dataset}. Images $x$ are obtained from BAM-FG \cite{aladin}, while style images $s$ are extracted from BAM (Behance Artistic Media Dataset) \cite{behance} and content images $y$ are parsed from \textit{Flickr}. BAM-FG is a dataset that contains 2.62M images grouped into 310K style-consistent groupings. Only a subset of stylized non-photorealistic images from BAM-FG with semantically rich content are considered for building our training triplets.

For the Generative Search experiment, a subset of 1M images from BAM is indexed and used for search while ensuring no intersection with the images used for training.

\textbf{Evaluation metrics.} We measure several properties of the style transfer quality in our experiments. LPIPS \cite{lpips} as a perceptual metric for the semantic similarity between the content and stylized image, SIFID \cite{singan} to measure style distribution similarity, and Chamfer distance to calculate colour similarity. We normalize Chamfer distance by the number of pixels in the image to maintain comparable values for any image resolution. We compute the MSE based on ALADIN embeddings to measure style similarity between the synthesized image and the input style and the MSE based on CLIP embeddings for quantifying its semantic similarity to the content input.

\subsection{Comparison to State-Of-The-Art Methods}
\label{seq:baselines}

\begin{figure}[t!]
    \centering
    \includegraphics[width=\linewidth,trim=0cm 0cm 0cm 0cm,clip]{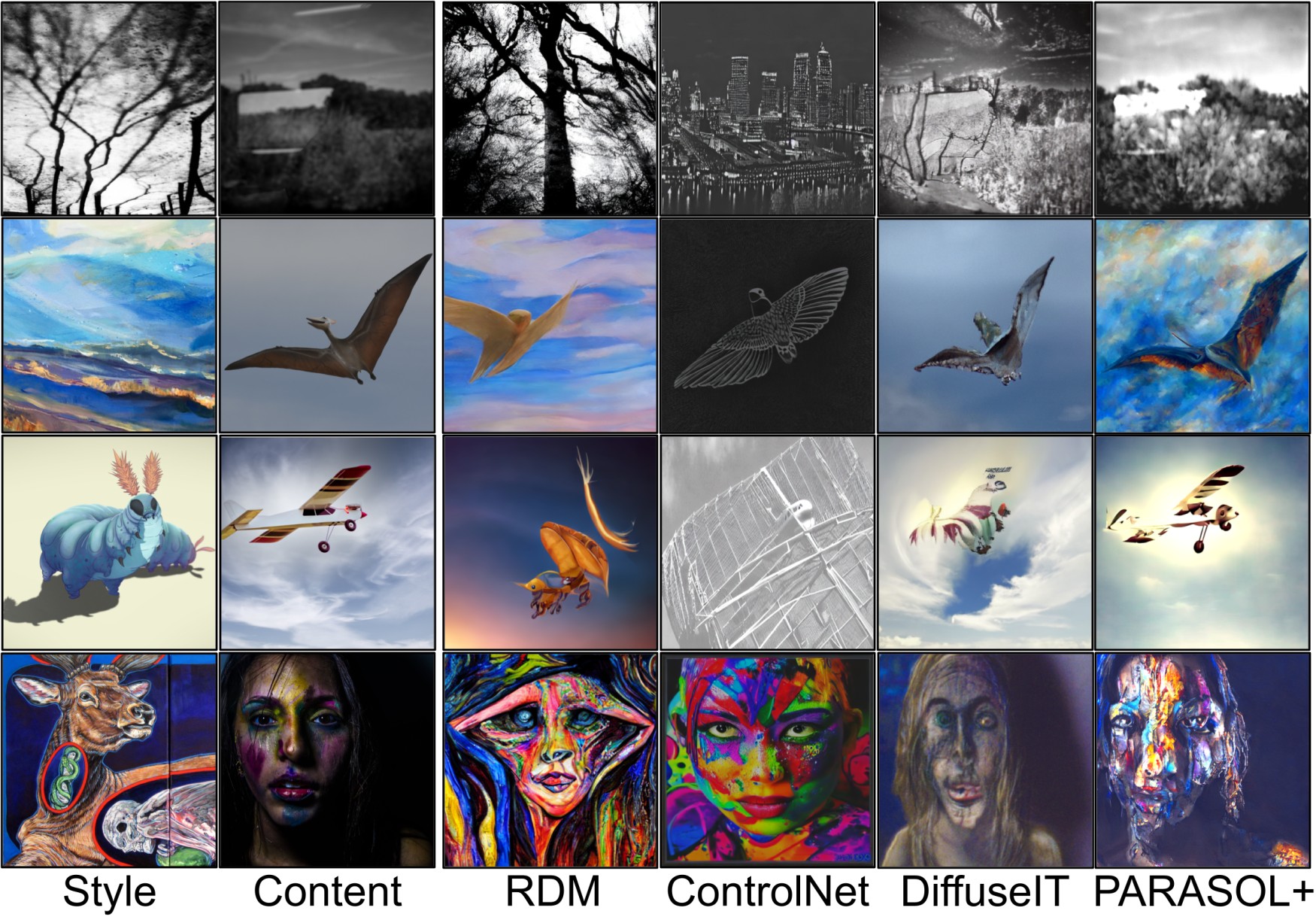}
    \caption{Comparison to Generative Multimodal Models (RDM \cite{rdm}, ControlNet \cite{controlnet}, DiffuseIT \cite{diffuseit}) and PARASOL+.} 
    \label{fig:baselines}
    \vspace{-5mm}
\end{figure}

\begin{figure*}[t!]
    \centering
    \includegraphics[width=\linewidth,trim=0cm 0cm 0cm 0cm,clip]{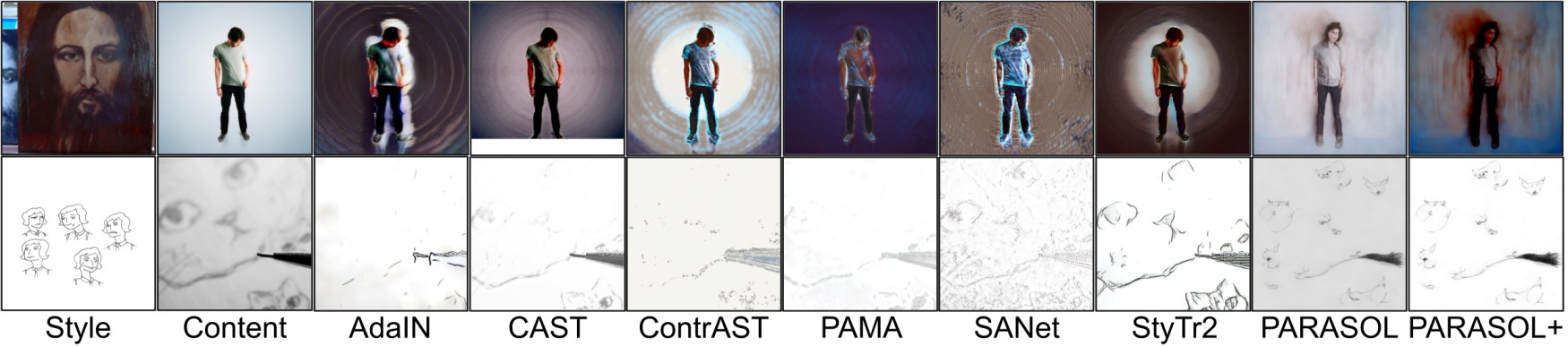}
    \vspace{-5mm}
    \caption{Comparison to Style Transfer Models (AdaIN \cite{adain}, CAST \cite{cast}, ContrAST \cite{contraAST}, PAMA \cite{pama}, SANet \cite{sanet}, StyTr2 \cite{deng2021stytr}), PARASOL and PARASOL+.}
    \label{fig:baselinesnst}
    \vspace{-5mm}
\end{figure*}

We separately compare to (i) generative multimodal diffusion-based models, (ii) NST models. For fairness, we fine-tune each of them on our training data.
\vspace{-4mm}
\paragraph{A) Comparison to Generative Multimodal Models} We compare to several SoTA models that can be conditioned on an example-based style.
\textit{RDM \cite{rdm}} share its backbone with PARASOL, the main difference being using CLIP for encoding all conditions. We fine-tune it on our training triplets.
\textit{ControlNet \cite{controlnet}} propose the training of a neural network for incorporating task-specific conditions into the generation process. Since it only accepts textual prompts as input, we extract captions from our content images using BLIP \cite{blip} and train the network to be conditioned by style images via our training triplets.
\textit{DiffuseIT \cite{diffuseit}} propose a diffusion-based image translation method guided by a semantic cue. Several losses ensure consistency in style and structure. Since no training code is provided, we resource to their publicly available model pretrained on ImageNet.

Visual examples of all generative baselines are shown in Fig. \ref{fig:baselines} and quantitative evaluations using the metrics described in Section \ref{seq:experimental_setup} are provided in Tab. \ref{tab:baselinesquant} (top). 

Our complete method produces the best SIFID and ALADIN-MSE scores proving to accurately transfer the specific input style. 
The measure of semantics similarity to the content input through CLIP-MSE and LPIPS is comparable to the other methods. Although our Chamfer score is improvable, we highlight the ability of our method for tuning the influence of style and structure. If style and colour similarity are a priority for the user, the different parameters in the model can be adapted for providing more stylistically similar images to the style input.

The optional post-processing step drastically improves Chamfer distances, and SIFID metrics. However, the LPIPS scores are worse. While this is not unexpected, it highlights a weakness with LPIPS, as a metric in measuring content similarity. The re-colouring steps change only the colours in the image, not modifying any content.

We undertake a user study using Amazon Mechanical Turk (AMT), to support our quantitative metrics. 
As show in Tab. \ref{tab:baselinesuser} (top), PARASOL was chosen as the preferred method in all categories, measured via majority consensus voting (3 out of 5 workers).

\begin{table}[t!]

\centering
\begin{adjustbox}{width=0.5\textwidth}

\begin{tabular}{lccccc}
\toprule
\textbf{Method}                           & \textbf{SIFID$\downarrow$} & \textbf{LPIPS$\downarrow$} & \textbf{ALADIN-MSE$\downarrow$} & \textbf{CLIP-MSE$\downarrow$} & \textbf{Chamfer$\downarrow$} \\ 
\cmidrule{1-6}
RDM \cite{rdm}        &      3.937          &       0.736                                                            &           3.945                      &      \textbf{12.792}                                                             &    1.431       \\
\cmidrule{1-6}
ControlNet \cite{controlnet} &      4.265             &                          0.757                                           &           5.148                        &           16.944                                                        &      3.612     \\ 
\cmidrule{1-6}
DiffuseIT \cite{diffuseit}          &  2.572    &      0.677                                                             &        4.460            &            15.159                                                       &    \textbf{0.184}   \\ 
\cmidrule{1-6}
PARASOL            &    2.994   &   \textbf{0.525}             &       4.054                                                      &          15.12                          &            1.847                     \\
\cmidrule{1-6}
PARASOL+           &    \textbf{2.293}   &   0.679             &       \textbf{3.891}                                                      &          14.999                          &                             0.340    \\

\cmidrule{1-6}\morecmidrules\cmidrule{1-6}
AdaIN \cite{adain}     &     2.290            &      0.631                                                         &        4.107                        &                   16.129                                               &  0.456   \\
\cmidrule{1-6}
CAST \cite{cast}     &     6.106            &      0.608                                                         &        4.633                        &                   16.732                                               &  0.188   \\
\cmidrule{1-6}
ContrAST \cite{contraAST}     &    2.791             &                           0.651                                    &       4.171                          &                   16.111                                               &     \textbf{0.126} \\
\cmidrule{1-6}
PAMA \cite{pama}     &     2.509            &                                     0.659                          &                         4.411          &                    16.194                                              &    0.287 \\
\cmidrule{1-6}
SANet \cite{sanet}     &    2.608             &                            0.671                                   &         4.200                       &                   16.187                                               &  0.301   \\
\cmidrule{1-6}
StyTr2 \cite{deng2021stytr}          &  \textbf{1.966}    &      0.588                                                             &        4.057            &            16.083                                                       &    0.535   \\ 
\cmidrule{1-6}

PARASOL            &    2.994   &   \textbf{0.525}             &       4.054                                                      &          15.12                          &            1.847                     \\
\cmidrule{1-6}
PARASOL+           &    2.293   &   0.679             &       \textbf{3.891}                                                      &          \textbf{14.999}                          &                             0.340    \\

\bottomrule
\end{tabular}
\end{adjustbox}
\vspace{-2mm}
\caption{Quantitative comparison of style (SIFID, ALADIN-MSE, Chamfer) and content (LPIPS, CLIP-MSE) metrics. (\textit{Top:} Generative Multimodal Models; \textit{Bottom:} Style Transfer Models) We include our optional post-processing as PARASOL+. Chamfer scores are scaled down $\times 10^{-3}$.} 
\label{tab:baselinesquant}
\vspace{-5mm}
\end{table}


\vspace{-4mm}
\paragraph{B) Comparison to Style Transfer Models} Since most diffusion-based models are text-based, we mostly resource to state of the art non-diffusion based models. We evaluate our method against five recent NST models (CAST \cite{cast}, PAMA \cite{pama}, SANet \cite{sanet}, ContraAST \cite{contraAST} and StyTr2 \cite{deng2021stytr}), as well as a traditional one (AdaIN \cite{adain}). Each of these methods is prompt-free and directly comparable to our method. Note that PARASOL is \textit{not} designed to be a style transfer model. However, we display in Tab. \ref{tab:baselinesquant} (bottom) how our model is able to preserve both style and content comparably to the SoTA style transfer models, which were specifically trained for texture-based stylization instead of generation. Additionally, the user studies in Tab. \ref{tab:baselinesuser} (bottom) show how users prefer PARASOL over the most popular NST SoTA models in all experiments. One benefit of PARASOL when compared to most NST models is that, by relaxing the constraint of preserving exact content, it can adapt to more complex styles while maintaining unchanged semantics (See Fig. \ref{fig:baselinesnst}).

\begin{table}[t!]

\centering
\begin{adjustbox}{width=0.5\textwidth}

\begin{tabular}{lccc}
\toprule
\textbf{Method}                           & \textbf{Pref. Overall} & \textbf{Pref. Style Fidelity} & \textbf{Pref. Content Fidelity} \\ 
\cmidrule{1-4}
RDM \cite{rdm}       &       17.60\%         &                                    18.40\%                                &         9.20\%          \\
\cmidrule{1-4}
ControlNet \cite{controlnet} &        1.20\%           &                                        0.00\%                             &           1.60\%     \\ 
\cmidrule{1-4}
DiffuseIT \cite{diffuseit}          &  29.20\%    &                                              34.80\%                     &                   27.20\%   \\ 
\cmidrule{1-4}

PARASOL            &     \textbf{52.00\%}               &                                   \textbf{46.80\%}                            &                      \textbf{62.00\%}     \\
\cmidrule{1-4}\morecmidrules\cmidrule{1-4}

CAST \cite{cast}       &   20.40\%             &   11.20\%                                                                &             33.60\%      \\
\cmidrule{1-4}
ContrAST \cite{contraAST} &     15.20\%              &              12.00\%                                                       &             10.40\%     \\ 
\cmidrule{1-4}
PAMA \cite{pama}          & 17.60\%     &          8.40\%                                                         &               8.40\%      \\ 
\cmidrule{1-4}
SANet \cite{sanet}          &  16.40\%    &                       9.60\%                                            &         7.20\%     \\ 
\cmidrule{1-4}
PARASOL            &       \textbf{30.40\%}             &            \textbf{58.80\%}                                                   &             \textbf{40.40\%}          \\

\bottomrule
\end{tabular}
\end{adjustbox}
\vspace{-2mm}
\caption{Evaluation of our method vs. different baselines based on AMT experiments. (\textit{Top:} Generative Multimodal Models; \textit{Bottom:} Style Transfer Models). Given a content image, a style image and a set of images generated using our method and different baselines, we conduct three separate experiments. In each experiment, workers are asked to choose their preferred image based on: 1) image quality, 2) style fidelity, and 3) content fidelity. For fairness, we compare to PARASOL without the post-processing step.}
\vspace{-5mm}

\label{tab:baselinesuser}

\end{table}


\vspace{-2mm}
\subsection{Ablation Study} \label{seq:ablation}
\vspace{-1mm}

Considering the pre-trained RDM conditioned on CLIP embeddings as our baseline, we justify the addition of each of our components through an ablation study (Tab. \ref{tab:ablation}).

\begin{figure}[t!]
    \centering
    \includegraphics[width=0.8\linewidth,trim=0cm 0cm 0cm 0cm,clip]{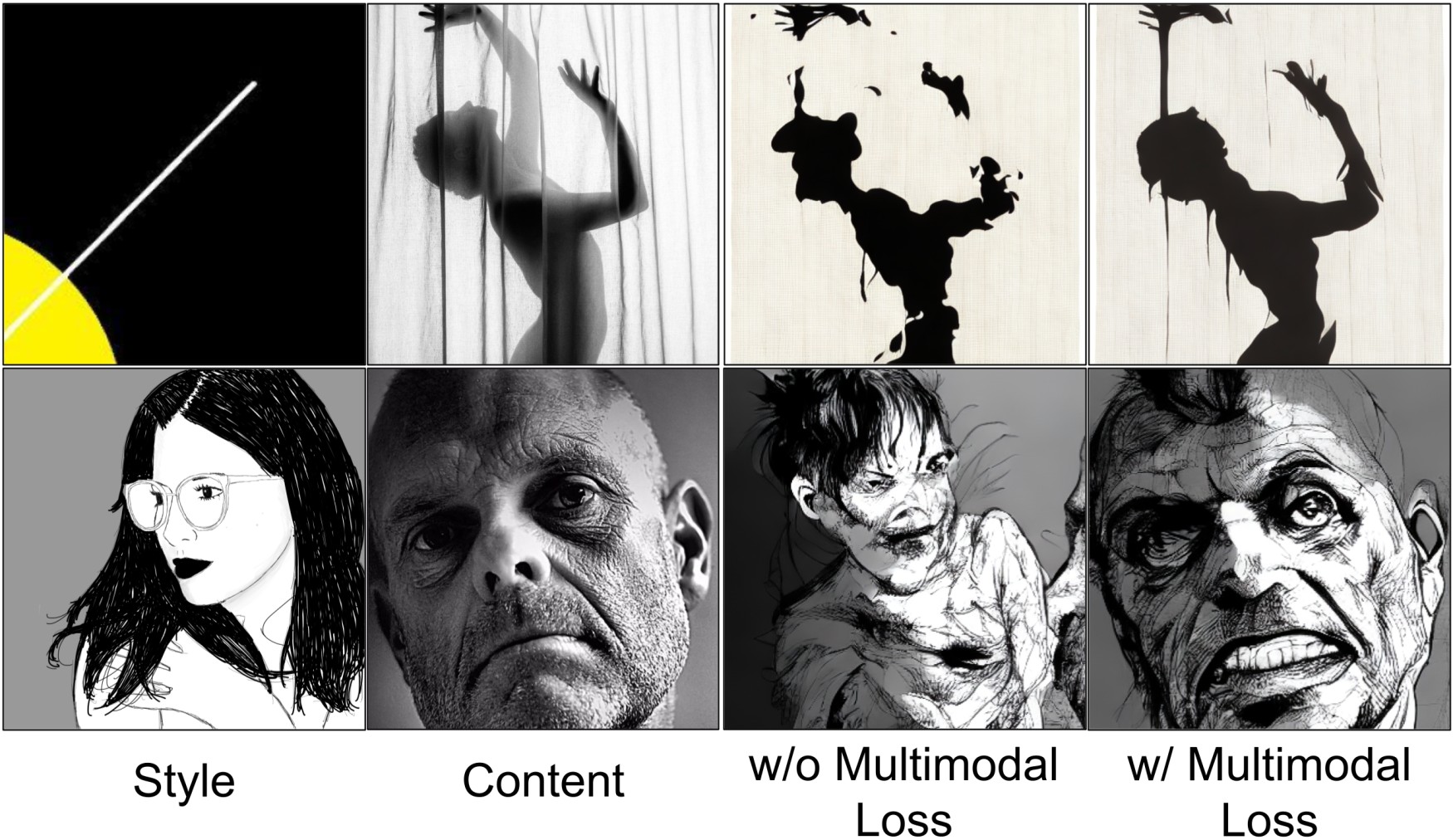}
    \vspace{-2mm}
    \caption{Effect of the multimodal loss. Adding the multimodal loss encourages the model to better combine the information from each modality when their descriptors are not fully disentangled.} 
    \label{fig:loss}
    \vspace{-2mm}
    
\end{figure}

Switching the style encoder from CLIP to ALADIN offers a lot of benefits in terms of disentanglement and fine-grained style information. However, even when fine-tuning the network on the new descriptors, it is not able to understand the nature of the new representations. Tab. \ref{tab:ablation} shows the substantial improvement of all metrics when the projector $M()$ is introduced, proving its crucial role in the pipeline. As illustrated in Fig. \ref{fig:loss}, the multimodal loss assists the network in further disentangling both modalities.

Despite some lower metrics when using inversion, we consider it beneficial overall, due to the added controllability, and improved style transfer metrics (SIFID, LPIPS), making PARASOL+ our best model overall.

\begin{table}[t!]

\centering
\begin{adjustbox}{width=\linewidth}

\begin{tabular}{lccccc}
\toprule
\textbf{Method}                           & \textbf{SIFID$\downarrow$} & \textbf{LPIPS$\downarrow$} & \textbf{ALADIN-MSE$\downarrow$} & \textbf{CLIP-MSE$\downarrow$} & \textbf{Chamfer$\downarrow$} \\ 
\cmidrule{1-6}
PARASOL -(I, L, M, Ft, A)          &    3.077               &                     0.749                                           &             4.312                        &          15.428                                                     &     \textbf{0.127}                                                                      \\ 
\cmidrule{1-6}
PARASOL -(I, L, M, Ft)        &       7.759         &                      0.813                                             &           5.540                          &          17.457                                                         &     1.184      \\

\cmidrule{1-6}
PARASOL -(I, L, M)           &   6.883   &                                           0.777                        &              4.356          &                                       15.887                            &     0.788  \\ 
\cmidrule{1-6}
PARASOL -(I, L)            &     4.269               &      0.748                                                         &        3.573                            &                        14.740                                     &    0.174    \\
\cmidrule{1-6}
PARASOL -(I)            &     4.329               &      0.747                                                        &        \textbf{3.564}                            &                        \textbf{14.714}                                     &    0.155    \\

\cmidrule{1-6}
PARASOL            &           2.994         &    \textbf{0.525}                                                           &      4.054                            &   15.12                                                          &   1.847     \\
\cmidrule{1-6}
PARASOL+            &     \textbf{2.293}               &              0.679                                                 &           3.891                        &     14.999                                                        &   0.340     \\

\bottomrule
\end{tabular}
\end{adjustbox}
\vspace{-2mm}
\caption{Quantitative evaluation metrics for different ablations of our final model. Considering pre-trained RDM as the baseline, this study justifies: 1) The use of a parametric style encoder (A); 2) The need to fine-tune the model (Ft); 3) The addition of a projector network for the style embedding (M); 4) The effect of modality-specific losses (L); 5) The design choice of inverting the diffusion process (I); 6) The use of the optional post-processing step (PARASOL+).}
\vspace{-2mm}
\label{tab:ablation}
\end{table}


\vspace{-2mm}
\subsection{Controllability Experiments}
\label{seq:controllability}
\vspace{-1mm}

We demonstrate several ways PARASOL can be used to exert control over the image synthesis process.

\vspace{-5mm}
\paragraph{A) Disentangled Style and Content}

We propose different ways of controlling the influence of each input modality. 

\begin{figure}[t!]
    \centering
    \includegraphics[width=\linewidth,trim=0cm 0cm 0cm 0cm,clip]{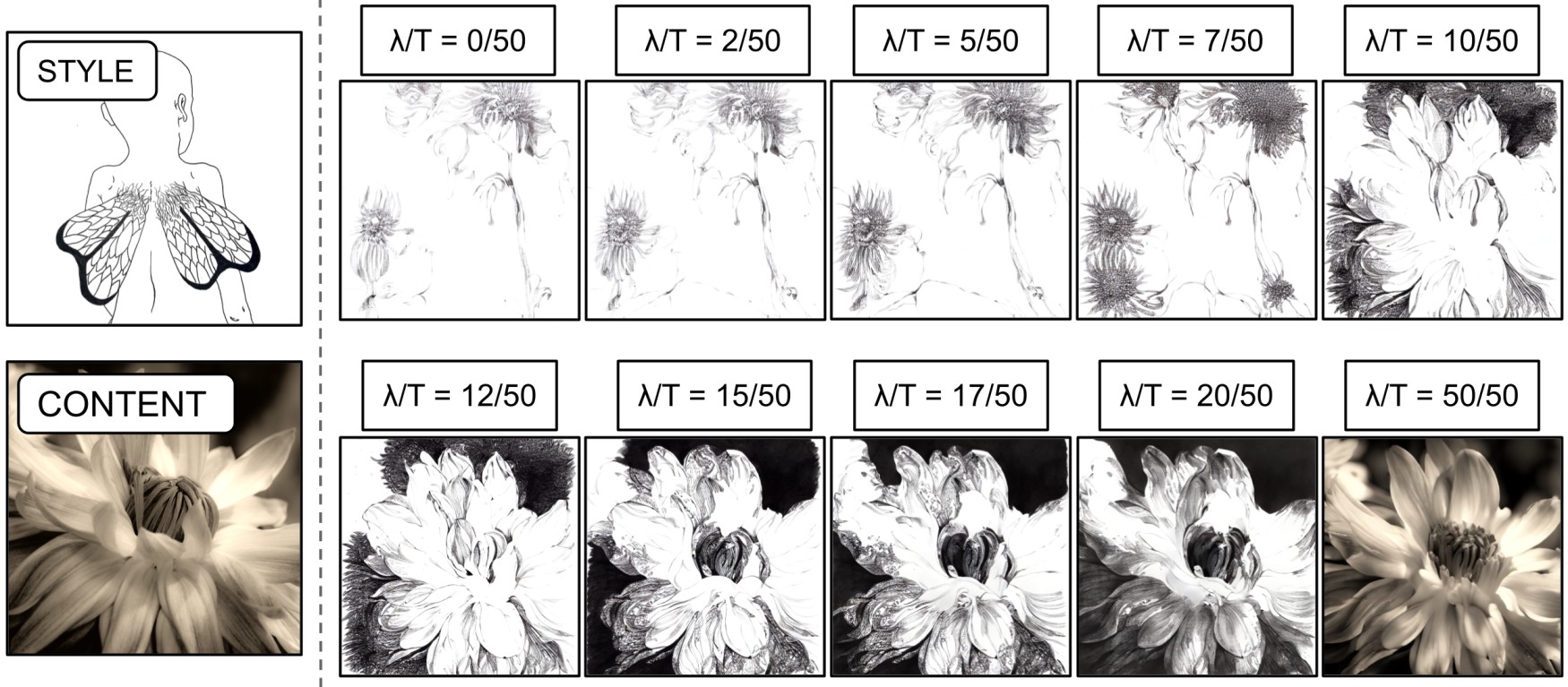}
    \caption{Effect of $\lambda$. 
    Higher $\lambda$ values lead to more content preservation, while lower values further encourage style transfer.}
    \vspace{-5mm}
    \label{fig:inversioncontrol}
\end{figure}

\textit{Via inversion:} $\lambda$ enables choosing at which step of the inversion process the style condition is changed (Fig. \ref{fig:inversioncontrol}). 

\begin{figure}[t!]
    \centering
    \includegraphics[width=0.95\linewidth,trim=0cm 0cm 0cm 0cm,clip]{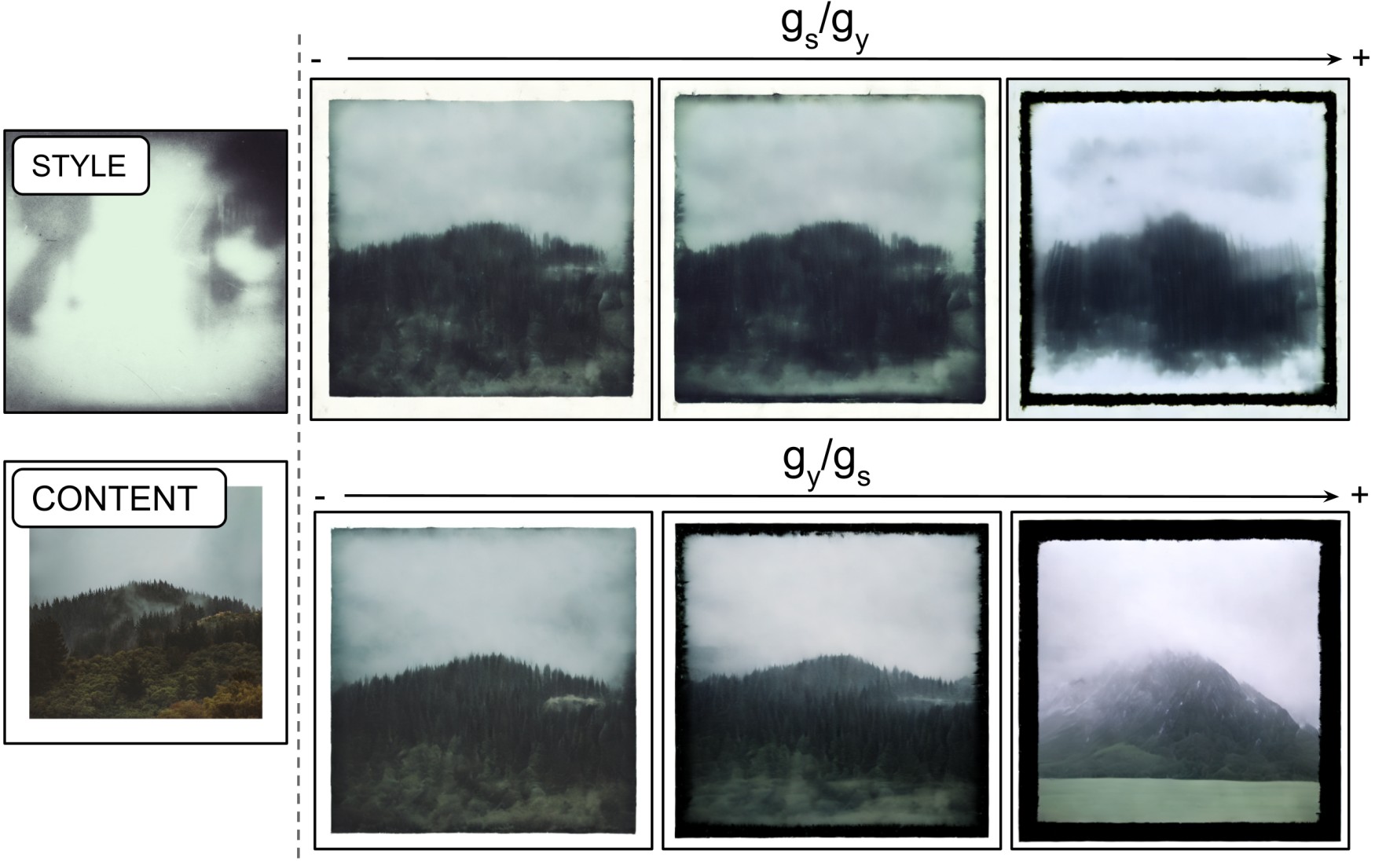}
    \caption{Effect of $g_s$ and $g_y$.  \textit{Top:} as the value of $g_s$ increases, so does the influence of the style. \textit{bottom:} $g_y$ is increased, offering a more notable semantic influence from the content cue.}
    \halfsqueezeup
    \label{fig:controlclassifierfree}
\end{figure}

\textit{Via classifier-free guidance parameters:} The values $g_s$ and $g_y$ (Eq. \ref{eq:gamma}) determine how much weight each condition has in the image generation process. (Fig. \ref{fig:controlclassifierfree}).

\squeezeup
\paragraph{B) Textual Captions as Conditioning Inputs}
\label{seq:inputcaptions}
Leveraging the multimodal properties of CLIP, content cues can be provided as either captions or images at sampling time. When a textual prompt is provided, our diffusion model is used for generating an image conditioned on its CLIP descriptor and an empty style descriptor. Considering this generated image as $y$, the sampling procedure continues as in Section \ref{seq:sampling}.

\begin{figure}[t!]

    \centering
    \includegraphics[width=0.8\linewidth,trim=0cm 0cm 0cm 0cm,clip]{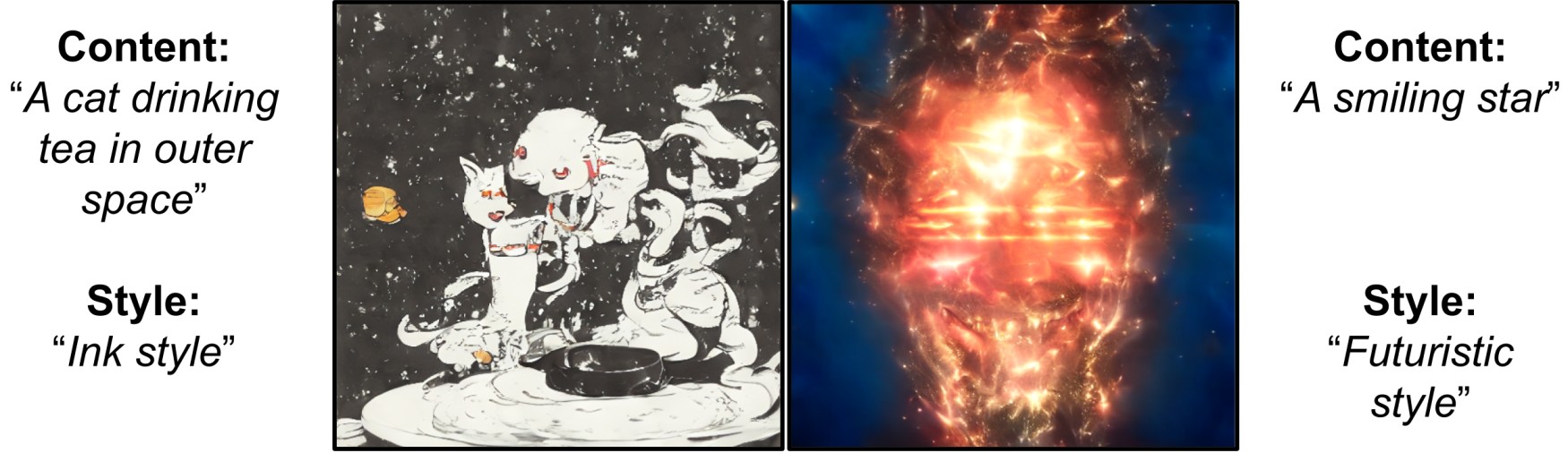}
    \caption{Images generated from text prompts for both modalities. PARASOL accepts both visual and textual inputs for better transferring the user's intent.}
    \vspace{-5mm}
    \label{fig:input_text}
\end{figure}

The style cue can also be provided in textual format. The prompt can be encoded using CLIP and projected to a joint space through a pre-trained projector network \cite{stylebabel}. By indexing all images in the ``Style Database" $\mathcal{S}$ through ALADIN and projecting them to the joint feature space, a similar style image can be retrieved and used as input $s$ to the sampling pipeline (Fig. \ref{fig:input_text}).

\begin{figure}[t!]
    \centering
    \includegraphics[width=\linewidth,trim=0cm 0cm 0cm 0cm,clip]{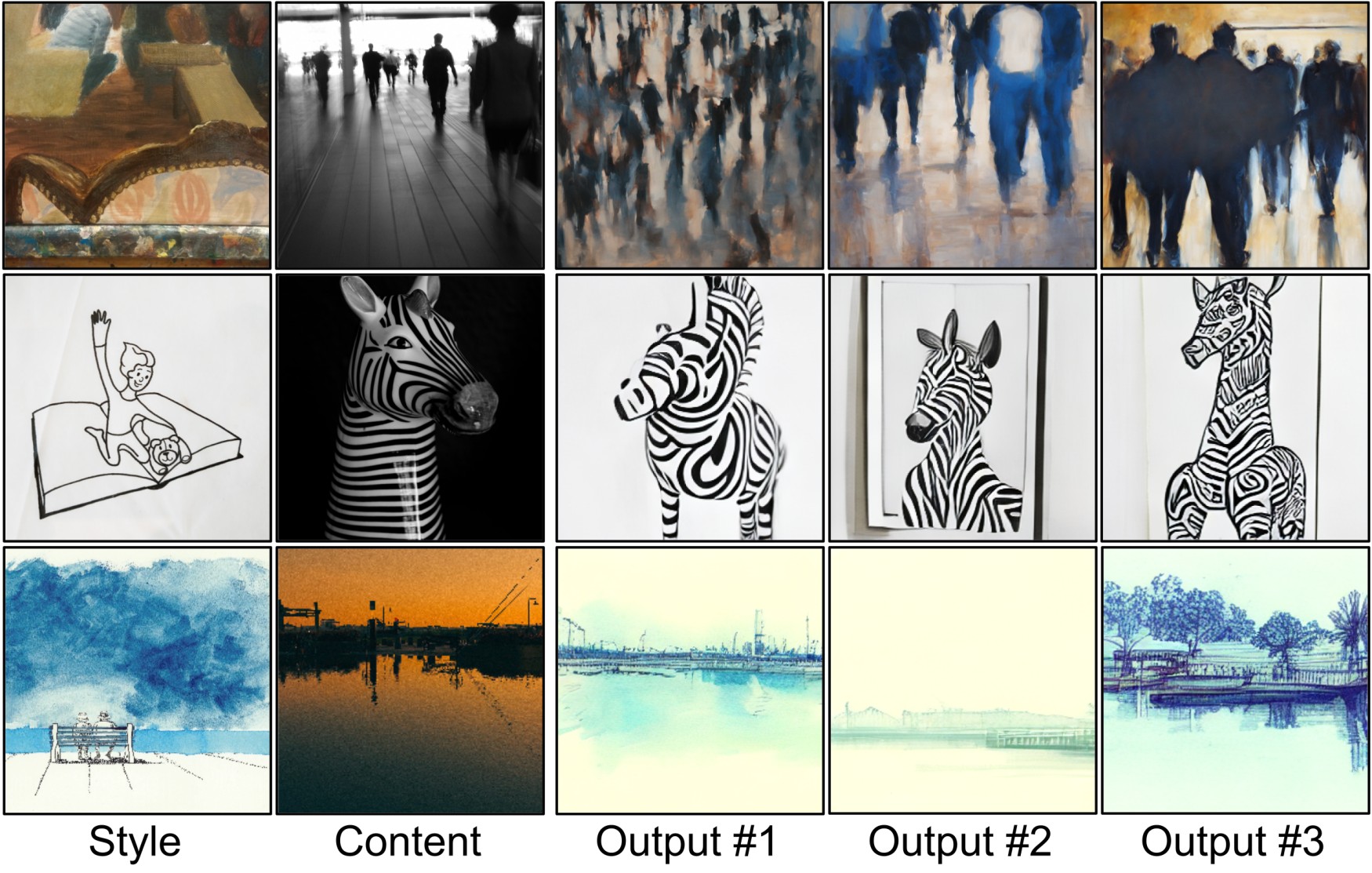}
    \vspace{-5mm}
    \caption{Diversity of fine-grained content. PARASOL can be used for synthesizing images with same semantics and fine-grained style yet diverse fine-grained content.}
    \vspace{-2mm}
    \label{fig:inversionfromgenerated}
\end{figure}

\vspace{-3mm}
\paragraph{C) Content Diversity with Consistent Semantics}
\label{seq:diversity}
Although PARASOL offers the option of preserving the fine-grained content details in $y$, it can also be used for synthesizing images with consistent semantics and style yet diverse details and image layouts (Fig. \ref{fig:inversionfromgenerated}). Following a similar procedure to that in Section \ref{seq:inputcaptions} (B), an image can first be generated from $y$ and used to guide the inversion process for transferring the semantics of $y$ with new fine-grained content details.

\subsection{Applications}
\label{seq:applications}

\begin{figure}[t!]
    \centering
    \includegraphics[width=0.9\linewidth,trim=0cm 0cm 0cm 0cm,clip]{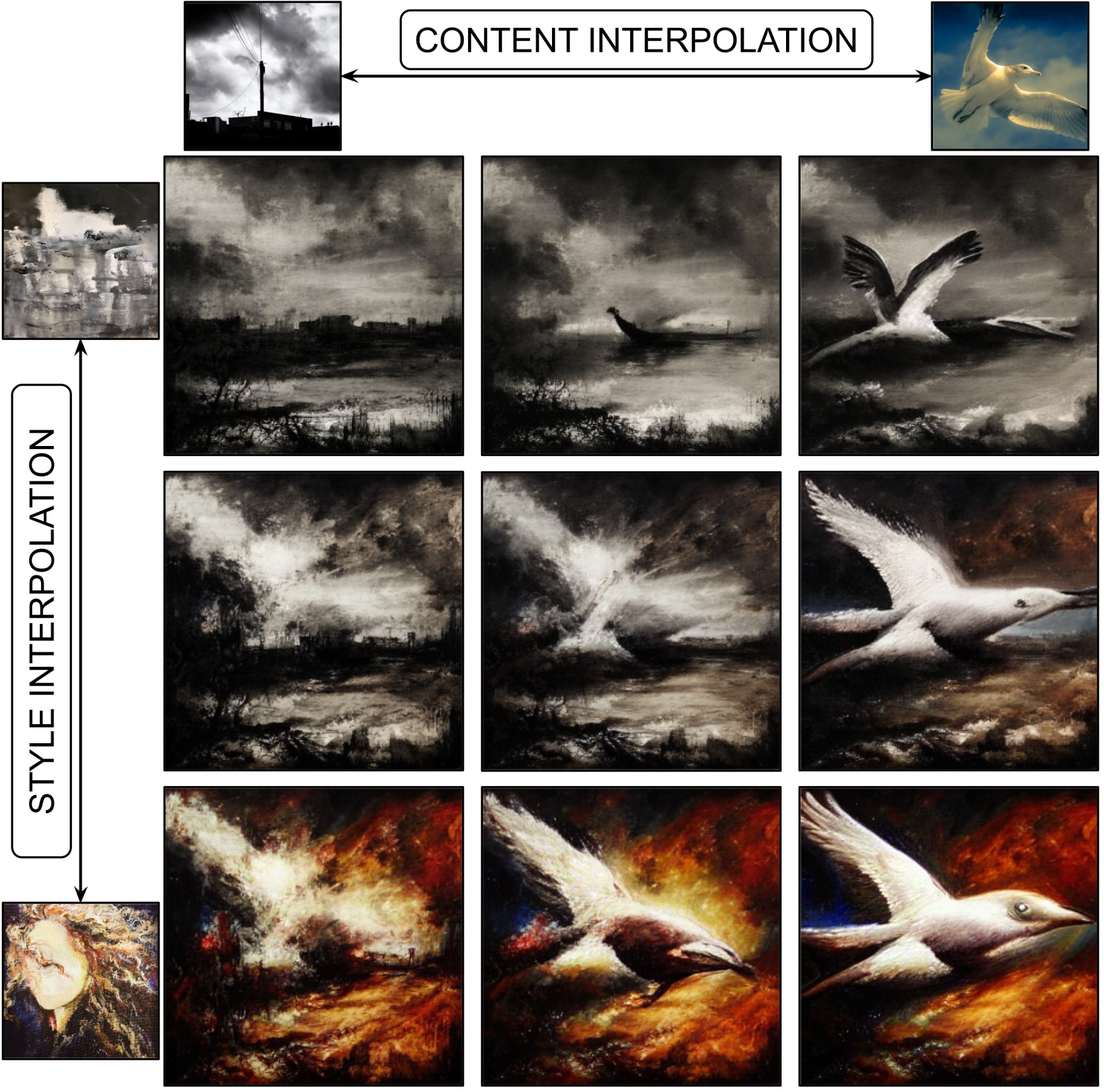}
    \vspace{-2mm}
    \caption{Style and content interpolation. Images generated by interpolating two styles and two content images.}
    \vspace{-6mm}
    \label{fig:sphericalinterpolation}
\end{figure}

\paragraph{A) Content and Style Interpolation}
\label{seq:interp}
PARASOL can synthesize images by interpolating different styles and/or contents, unlocking the potential of generating a wider range of creative images (Fig. \ref{fig:sphericalinterpolation}, Fig. 1). 
For demonstrating this capacity, a crowd-source AMT evaluation is performed (Tab. \ref{tab:interpolationuserstudy}), positioning PARASOL as the preferred method by users in terms of both content and style interpolation. 

\begin{table}[t!]

\centering
\begin{adjustbox}{width=0.9\linewidth}
    \begin{tabular}{lccc}
        \toprule
        \textbf{User Study}                           & \textbf{RDM} & \textbf{DiffuseIT} & \textbf{PARASOL} \\ 
        \cmidrule{1-4}
        Pref. Content Interpolation          &    12.037\%               &                     43.518\%                                           &             \textbf{44.444\%}                \\ 
        \cmidrule{1-4}
        Pref. Style Interpolation        &       16.808\%        &                      31.932\%                                             &          \textbf{51.260\%}      \\
    
        \bottomrule
    \end{tabular}
\end{adjustbox}
\vspace{-2mm}
\caption{AMT user evaluation of content and style interpolations, comparing PARASOL with RDM \cite{rdm} and DiffuseIT \cite{diffuseit}. Model preference is measured via majority consensus voting (3 out of 5).}
\vspace{-2mm}
\label{tab:interpolationuserstudy}
\end{table}

\begin{figure}[t!]
    \centering
    \includegraphics[width=0.8\linewidth,trim=0cm 0cm 0cm 0cm,clip]{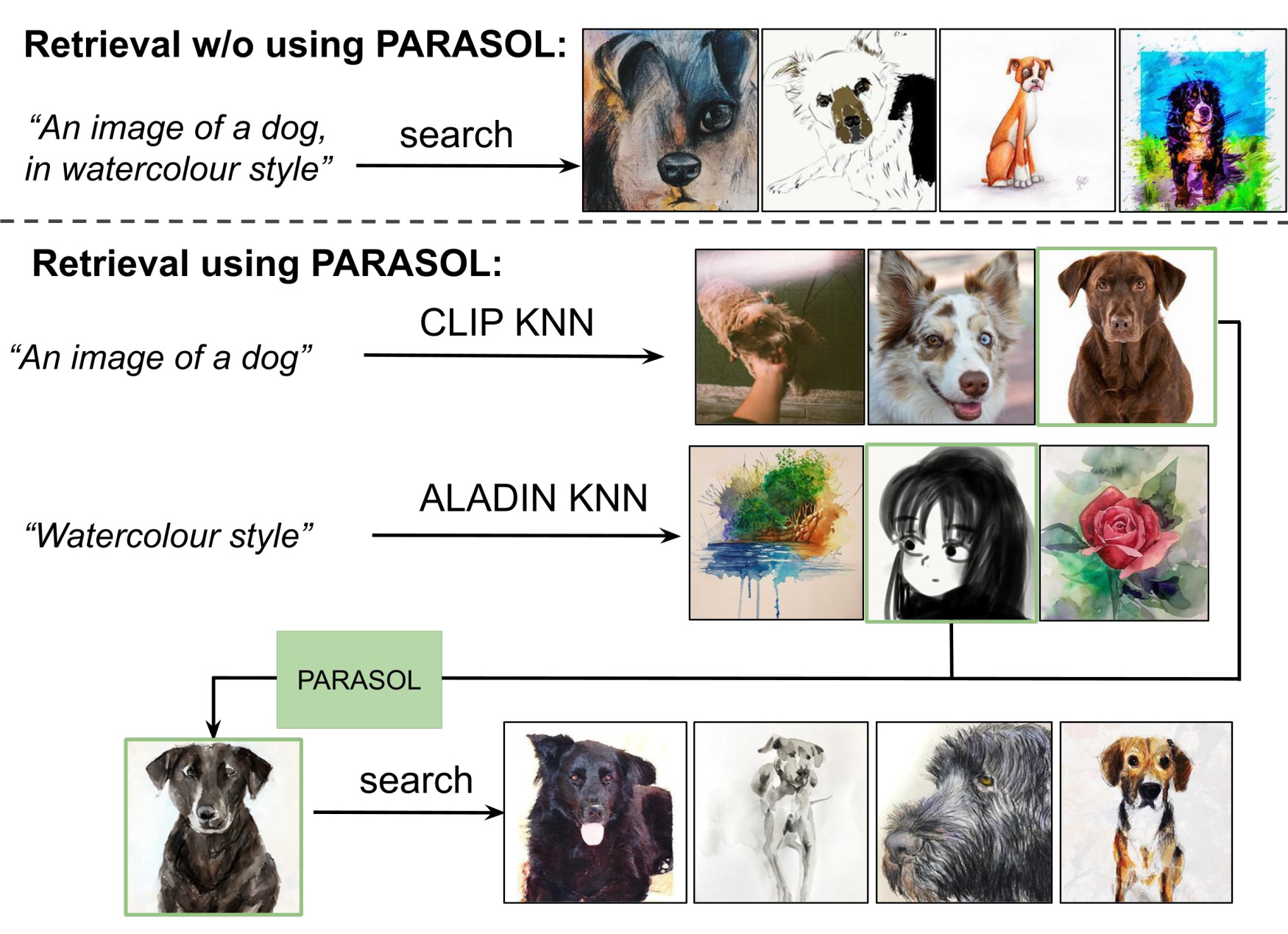}
    \caption{Generative Search. Our model can be used for aiding in image retrieval process (See Section \ref{seq:generativesearch} (B)).}
    \vspace{-2mm}
    \label{fig:generativesearch}
\end{figure}

\vspace{-5mm}
\paragraph{B) Generative Visual Search}
\label{seq:generativesearch}
PARASOL can be applied for the task of Generative Search, where search results are not constrained to a static corpus but fused with generation to yield images that more closely match a user's search intent.  Assuming an initial query in textual form, we use pre-existing text encoders (per Section \ref{seq:inputcaptions} (B)) for CLIP and ALADIN to retrieve two sets of images that match their intent semantically and stylistically (Fig. \ref{fig:generativesearch}, middle).  The user picks images from each set, each of which may be interactively weighted to reflect their relevance to the users' intent.  Using the fine-grained interpolation method of Section \ref{seq:interp} (A),  PARASOL is then used to generate an image which may either be accepted by the user or used as a further basis for semantic and style search.  PARASOL thus provides a fine-grained way to disambiguate a single text prompt by expressing a pair of modalities and can help surface images beyond those that would be available in a static text corpus.

\begin{figure}[t!]
    \centering
    \includegraphics[width=0.7\linewidth,trim=0cm 0cm 0cm 0cm,clip]{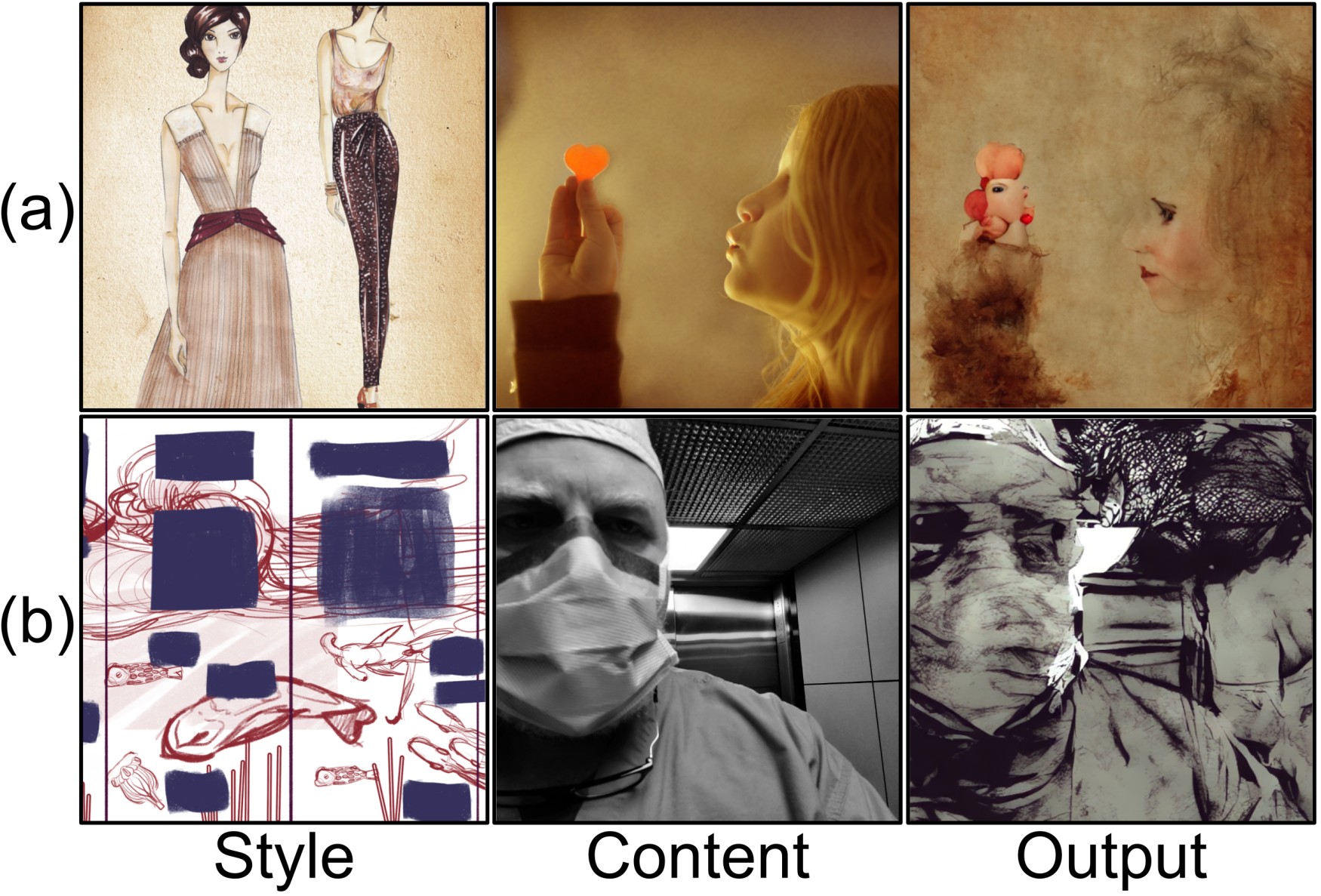}
    \vspace{-2mm}
    \caption{Failure cases - more details in Section \ref{seq:limitations}.}
    \vspace{-6mm}
    \label{fig:limitations}
\end{figure}


%% file: sec/4_conclusions.tex
\vspace{-3mm}
\section{Conclusion}
\vspace{-2mm}

We introduced PARASOL, a method that leverages parametric style embeddings for multimodal image synthesis with fine-grained parametric control over style. We show PARASOL can be applied to text, image or embedding conditioned generation enabling  disentangled control over style and content at inference time. We also propose a novel method for obtaining paired training data leveraging cross-modal search.  We show a use case in generative search, providing an image that can be used as query for a more fine-grained search. 

\label{seq:limitations}
\textbf{Limitations} While the use of modality-specific encoders offers numerous advantages in attribute disentanglement and parametric control, facilitating interpolation and search, certain styles can still present challenges due to ambiguity regarding which features should be considered as part of the content or style (Fig. \ref{fig:limitations}a). Additionally, addressing challenging contents such as faces (Fig. \ref{fig:limitations} b) often necessitates specific additional training. Future research could explore alternative modalities (i.e. sketches or segmentation maps) to target regions locally with fine-grained cues \cite{scenecomposer}.


\vspace{-3mm}
\section{Acknowledgements}
\vspace{-1mm}

This work was funded in part by DECaDE -- the UKRI Centre for Decentralized Creative Economy -- under the EPSRC grant EP/T022485/1.

%% file: sec/X_suppl.tex
\clearpage
\setcounter{page}{1}
\setcounter{section}{0}
\maketitlesupplementary

\section{Training Supervision Data}

\begin{figure*}[t]
    \centering
    \includegraphics[width=\textwidth]{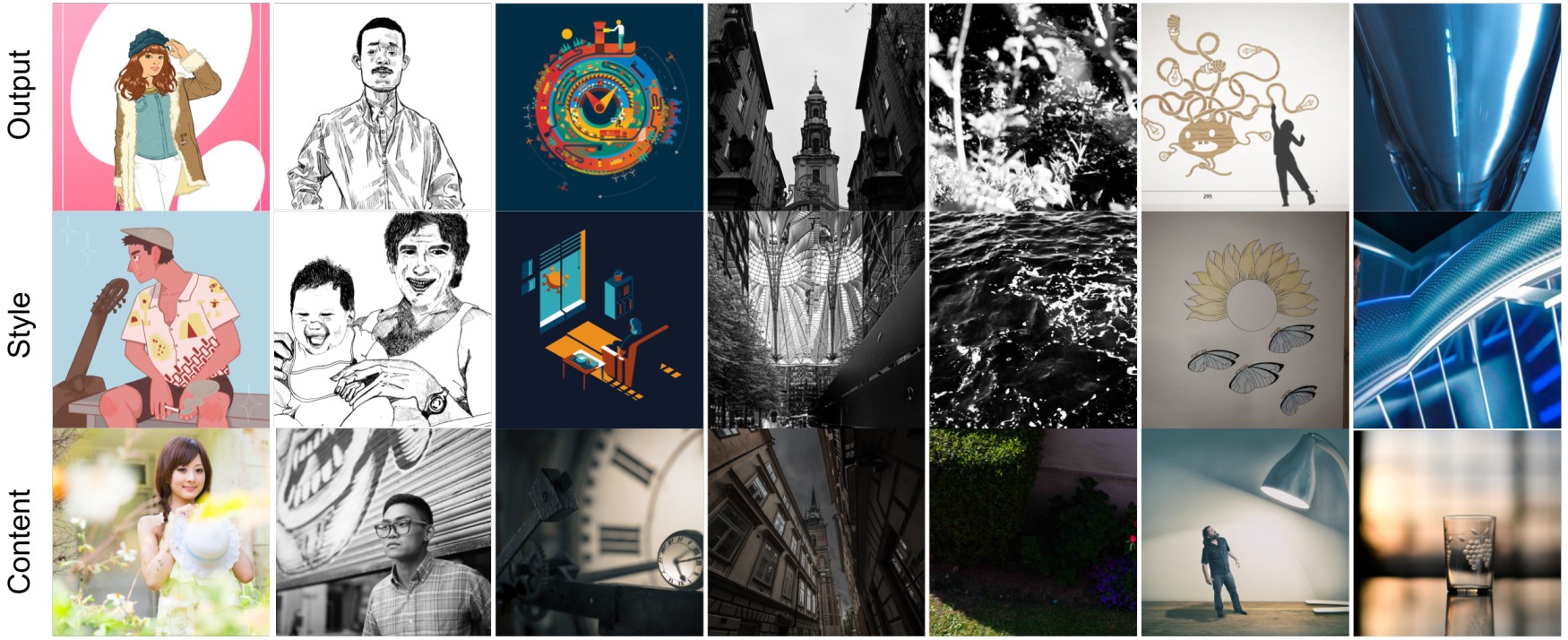}
    \caption{Visualization of triplets used for training our model. These triplets contain: output image \textit{x}, similar image in terms of style \textit{s} and similar image in terms of content \textit{y}. During training, the method learns to reconstruct $x$ by conditioning on $y$ and $s$. The complete set of training triplets will be released as a second contribution.} 
    \label{fig:dataset}
\end{figure*}

In order to obtain suitable data for training our model, a set of 500k triplets (output image \textit{x}, content image \textit{y}, style image \textit{s}) is obtained through cross-modal search (See Section \ref{seq:dataset} for details). The process in which these images are obtained ensures a certain disentanglement between content in $x$ and $s$ and between style in $x$ and $y$. These triplets are essential for the training of PARASOL and assist in disentangling the two attributes in which the network is conditioned. Fig. \ref{fig:dataset} shows a few examples of such triplets. 

 \section{Baselines Comparison}
 \label{seq:baselines_suppmat}

Quantitative and qualitative evaluations are provided in the main paper (Section \ref{seq:baselines}) comparing PARASOL to several image generation and neural style transfer (NST) methods.



The evaluations show how RDM \cite{rdm} and ControlNet \cite{controlnet} are the methods that less accurately keep the fine-grained style and control details. RDM encodes both conditions using the same kind of encoding, without encouraging any disentanglement, leading to confusion of the network in which attributes should be transferred from each input condition. For the comparison, ControlNet was trained following the author's indications and using our set of triplets as training data. It only accepts content information given in textual format, so we use automatically generated captions from each content image $y$ using BLIP \cite{blip}. Thus, the method was trained by feeding the generated captions as input and the style images as a conditioning that needs to be learnt. Their paper shows the method is able to learn a wide range of conditioning signals including sketches, segmentation maps and edgemaps. It doesn't, however, show any example in which the conditioning signal is not structure-based. Therefore, we hypothesize the architecture or parameters of ControlNet might not be suitable for successfully accommodating a condition such as style.

 \subsection{Comparison to eDiff-I}
 
 eDiff-I \cite{ediffi} is a diffusion-based method that generates images from text. It conditions the generation process on T5 text embeddings \cite{t5}, CLIP image embeddings and CLIP text embeddings \cite{CLIP}. The use of CLIP image embeddings allows extending the method for style transfer from a reference style image. 

\begin{figure}[t]
    \centering
    \includegraphics[width=\linewidth]{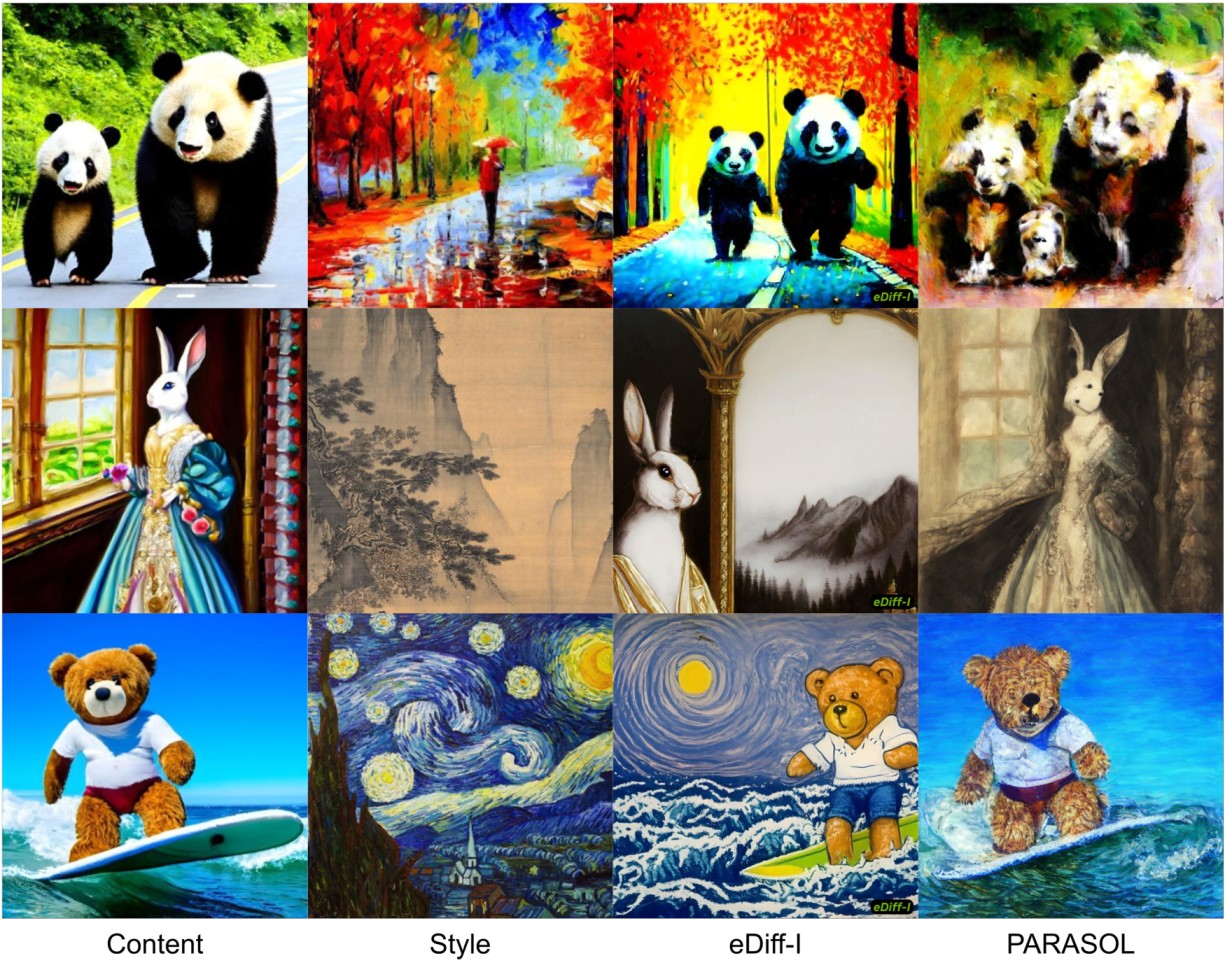}
    \caption{Comparison to eDiff-I \cite{ediffi}. PARASOL closely transfers the fine-grained style of the style image while keeping the fine-grained details and structure of the content image. The concept of style, however, slightly differs between both works. While eDiff-I understands "style" as a collection of colours and structure in which the information should be presented, PARASOL focuses on the type of artistic style (e.g. oil painting, illustration...) and its fine-grained details such as types of brush strokes, while also transferring the overall colour tonalities. }
    \label{fig:ediffi}
\end{figure}

 This work wasn't included in the previous baseline comparison due to lack of open source code or public pre-trained models. However, we show in Fig. \ref{fig:ediffi} a visual comparison to a set of synthetic images they provide. Those images were generated by eDiff-I from long descriptive captions and the displayed style images. They also provide unstylized images generated from the same descriptions without conditioning on any style. eDiff-I incorporates T5 text embeddings in their pipeline, allowing it to process much more complex text prompts than those that can be encoded through CLIP. Therefore, in our comparison, we consider as content input the provided unstylized synthetic images generated from the same prompts.


 \section{Amazon Mechanical Turk Experiments}

The results of 8 different Amazon Mechanical Turk (AMT) experiments were presented in the main paper, 6 of them comparing our method to different baselines in terms of style, content and overall preference (Section \ref{seq:baselines}) and 2 comparing style and content interpolations to DiffuseIT \cite{diffuseit} and RDM \cite{rdm} (Section \ref{seq:applications} (A)).

In particular, the instructions given to the workers in each task were the following:
\begin{itemize}
    \item Image generation preference in terms of style: \textit{"The photo has been transformed into the style of the artwork in multiple ways. Study the options and pick which most closely resembles the style of the artwork whilst also keeping the most structure detail in the photo."}
    \item Image generation preference in terms of content: \textit{"The photo has been transformed into the style of the artwork in multiple ways. Study the options and pick which keeps the best structure of the content, from details in the content image."}
    \item Image generation overall preference: \textit{"The photo has been transformed into the style of the artwork in multiple ways. Study the options and pick which most closely resembles the style of the artwork whilst also keeping the most structure detail in the photo."}
    \item Preferred method for content interpolation: \textit{"These images have been generated by interpolating Photo1 and Photo2 and transferring the style of the Artwork. Study the options and pick which set of images better display a smooth transition from content/semantics of Photo1 and Photo2 while maintaining good image quality and a consistent style similar to the Artwork."}
    \item Preferred method for style interpolation: \textit{"These images have been generated considering the structure in the Content image and an interpolation of styles from both Artworks. Study the options and pick which set of images better display a smooth transition from the style of Artwork1 to Artwork2 while maintaining good image quality and a consistent structure similar to the Content image."}
\end{itemize}

\begin{figure}[t]
    \centering
    \includegraphics[width=\linewidth]{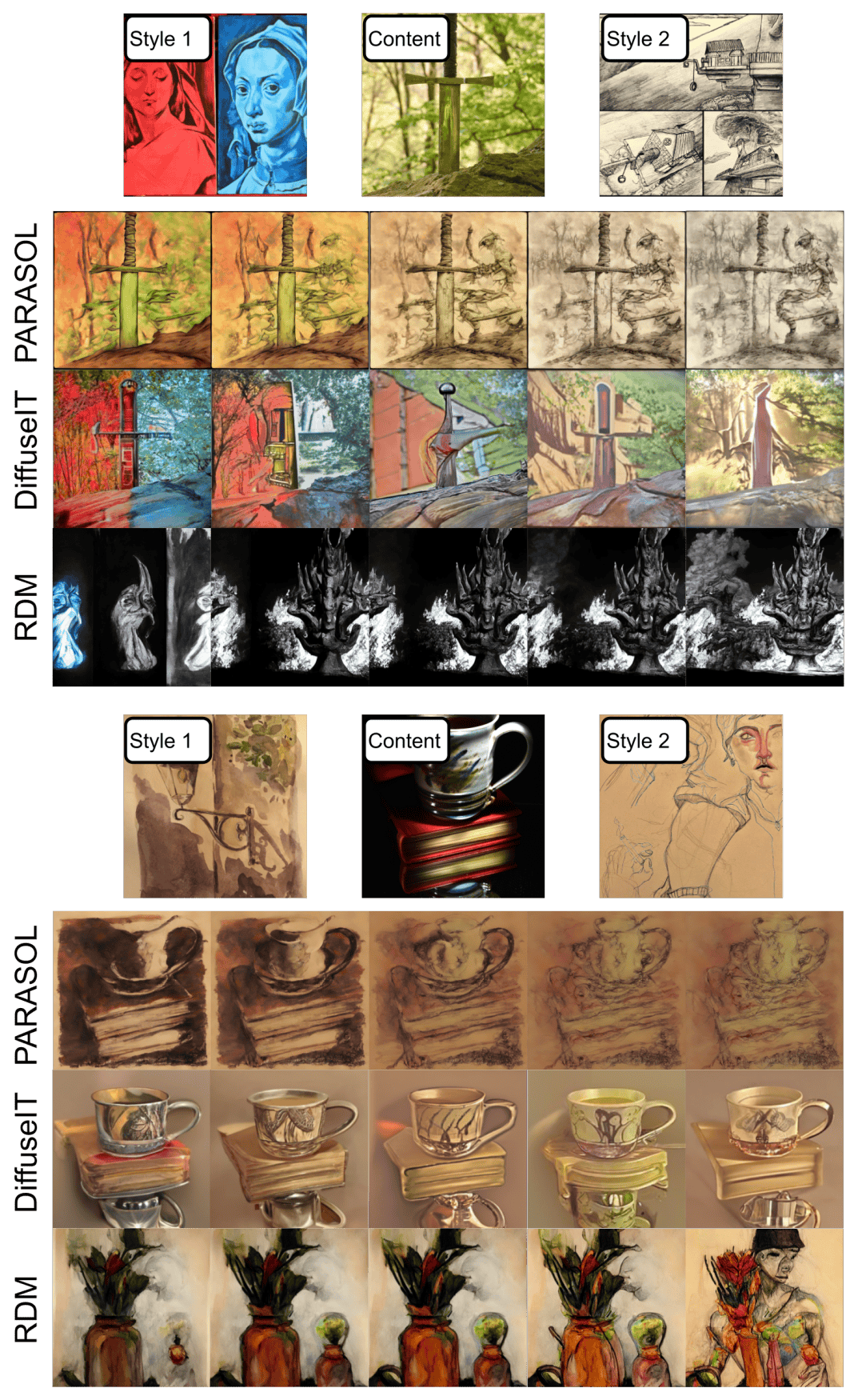}
    \caption{Baseline comparison for style interpolation. Visualization of images generated by PARASOL, DiffuseIT and RDM using interpolation between "Style 1" and "Style 2".}
    \label{fig:interpolation_baselines_style}
\end{figure}

\begin{figure}[t]
    \centering
    \includegraphics[width=0.9\linewidth]{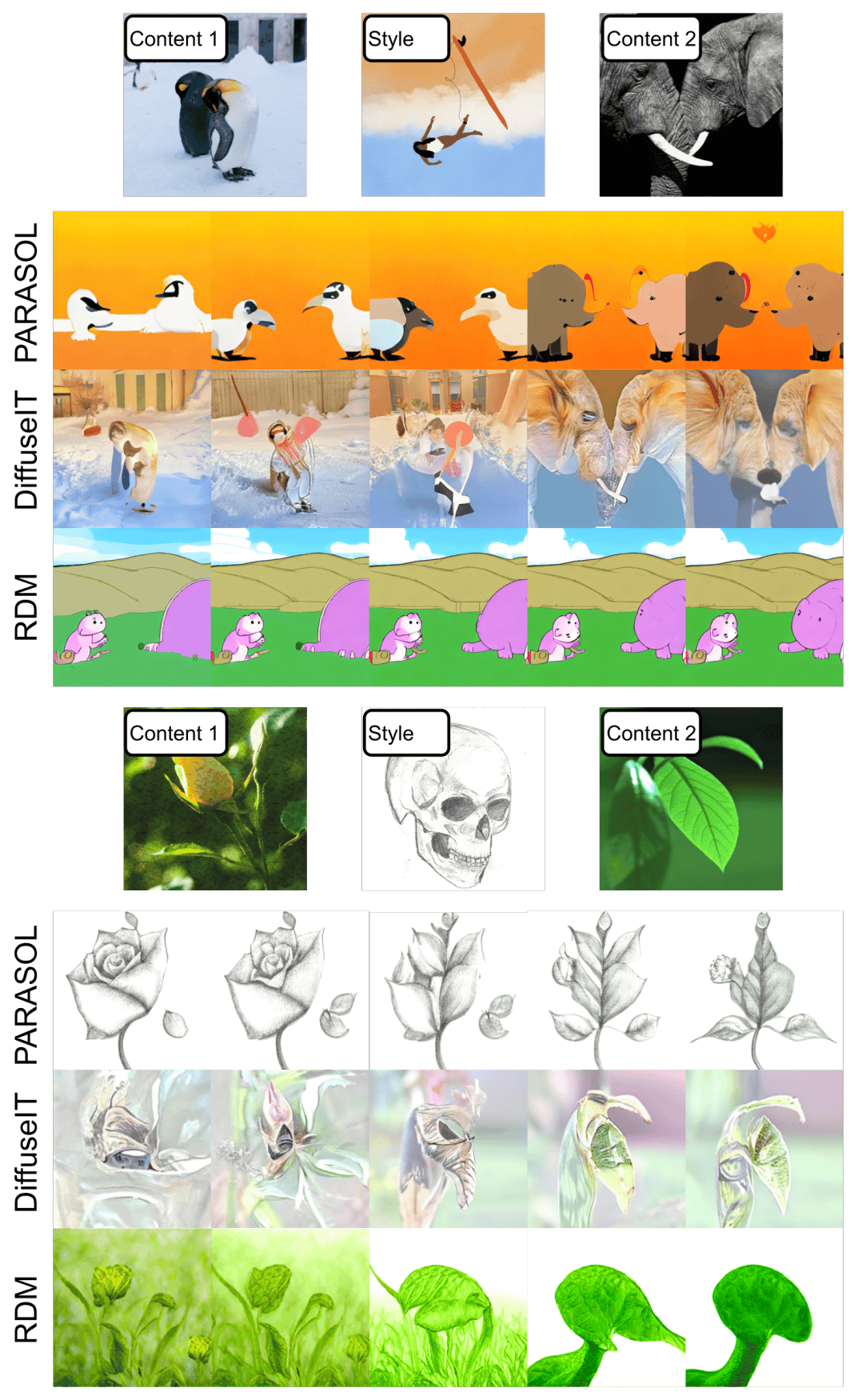}
    \caption{Baseline comparison for content interpolation. Visualization of images generated by PARASOL, DiffuseIT and RDM using interpolation between "Content 1" and "Content 2".}
    \label{fig:interpolation_baselines_content}
\end{figure}

The first three instructions were used for separately comparing PARASOL to image generation methods and NST ones. A few examples of the images shown to the users in the AMT interpolation experiments are shown in Fig. \ref{fig:interpolation_baselines_style}, \ref{fig:interpolation_baselines_content}.

\section{Generative Search}

\begin{figure}[t]
    \centering
    \includegraphics[width=\linewidth]{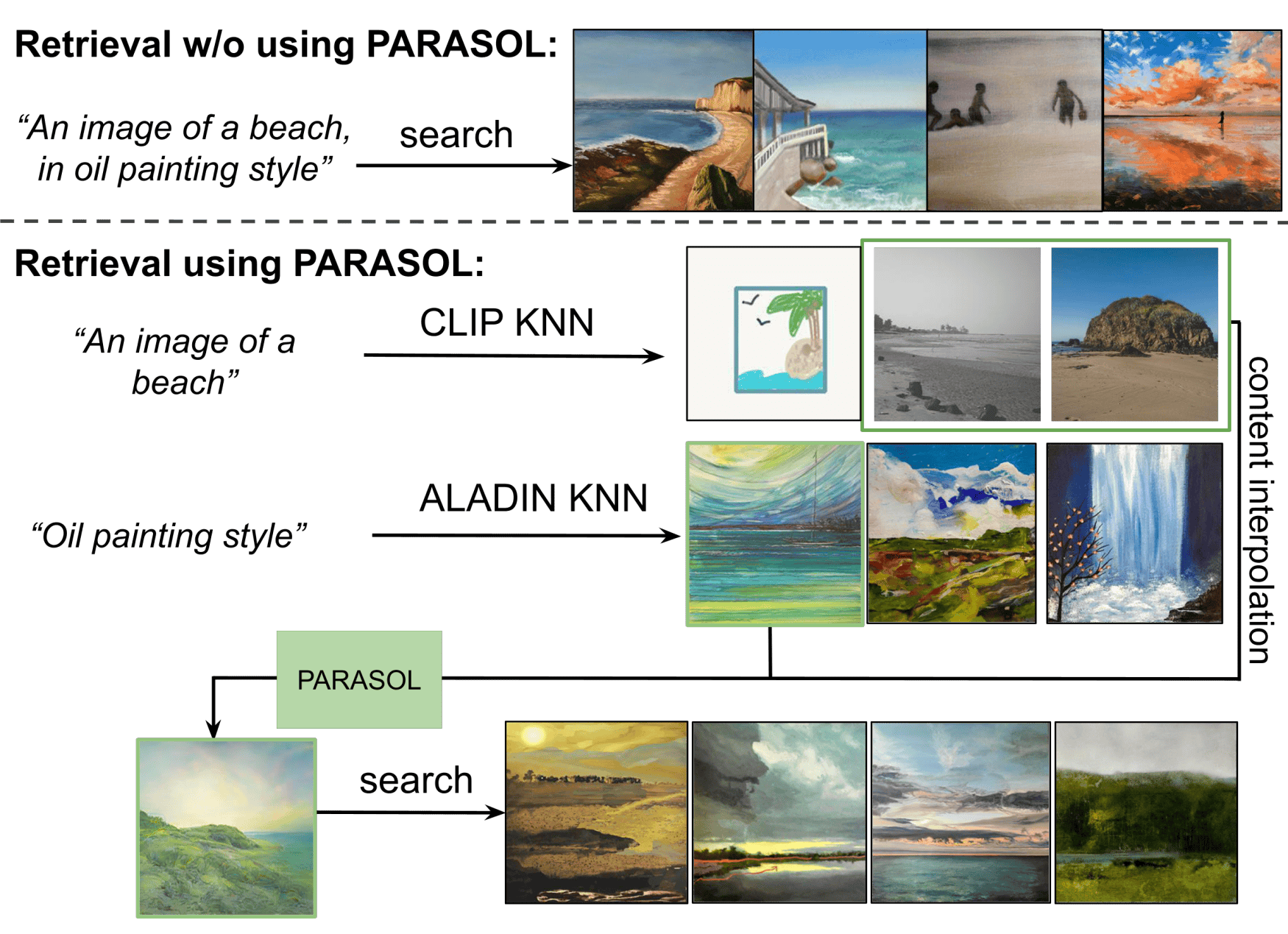}
    \caption{Use case for Generative Search. All PARASOL controllability tools, including interpolation capability, can be leveraged for obtaining a fine-grained query to refine the search and more closely match the user's intent.}
    \label{fig:gensearch}
\end{figure}

Briefly introduced in Section \ref{seq:applications} (B), generative search is presented as one of the main applications for our model. Fig. \ref{fig:gensearch} shows a different example of how PARASOL can be used for either refining the search with a more fine-grained query in terms of style and semantics or for generating a synthetic image that closely matches the user's intent. As depicted in this example, style and/or content properties of different existent images can be combined for a more fine-grained search.

 \section{Additional Visualization Examples}

We provide additional visualization examples for all experiments in Sections \ref{seq:controllability} and \ref{seq:applications} (A), as well as an additional controllability experiment. 


 \subsection{Images Generated from Textual Inputs}

\begin{figure}[t]
    \centering
    \includegraphics[width=0.7\linewidth]{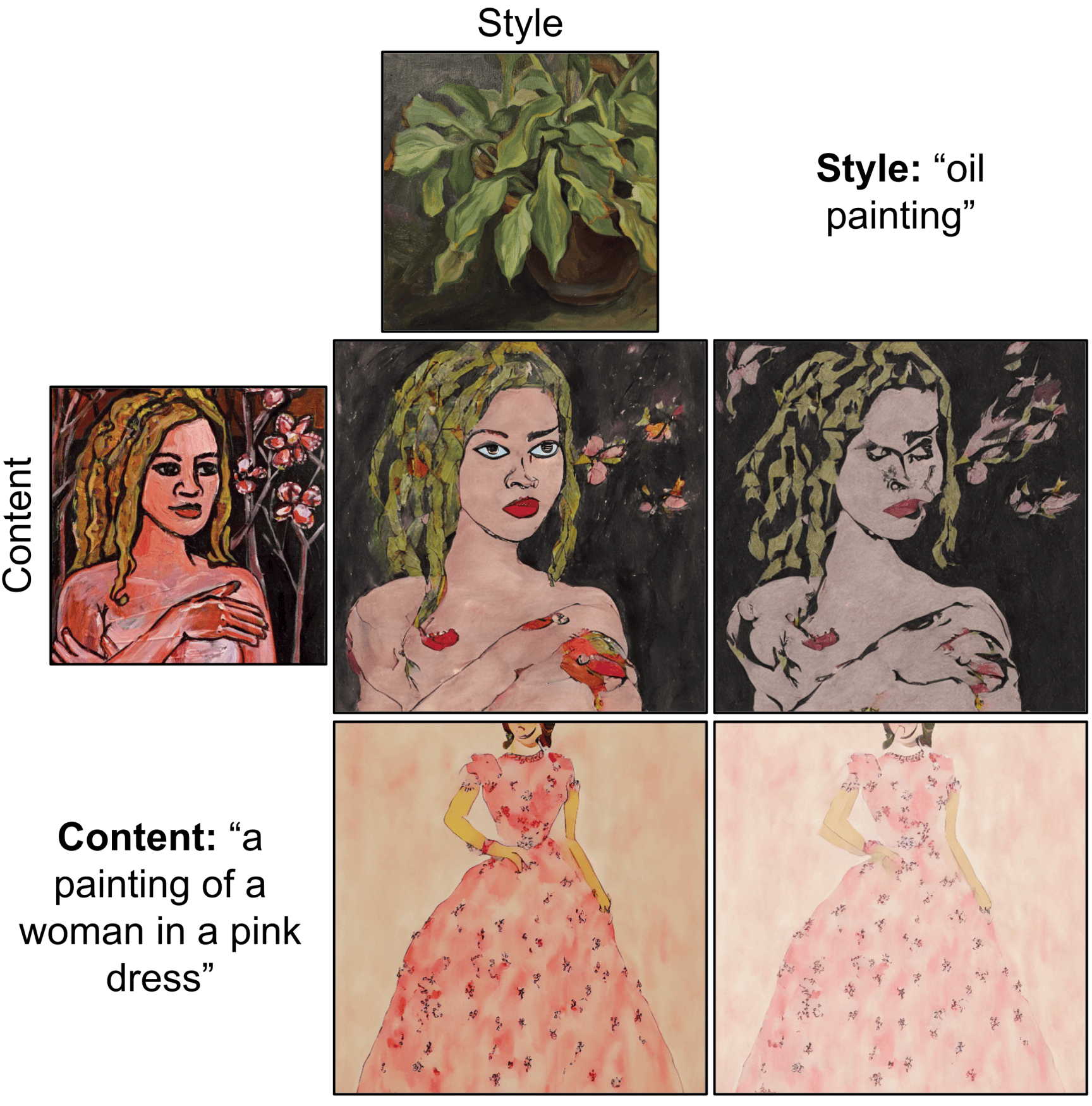}
    \caption{Text vs. Image to describe style and content conditions. While providing an image to describe intended content or style provides more fine-grained details, textual inputs allow useful descriptions and unlimited creativity.}
    \label{fig:text2img}
\end{figure}

An example of images generated with PARASOL using textual vs. image inputs for style and/or content is provided in Fig. \ref{fig:text2img}.

 \subsection{Images Generated with Different Lambda Values}

\begin{figure*}[t]
    \centering
    \includegraphics[width=\textwidth]{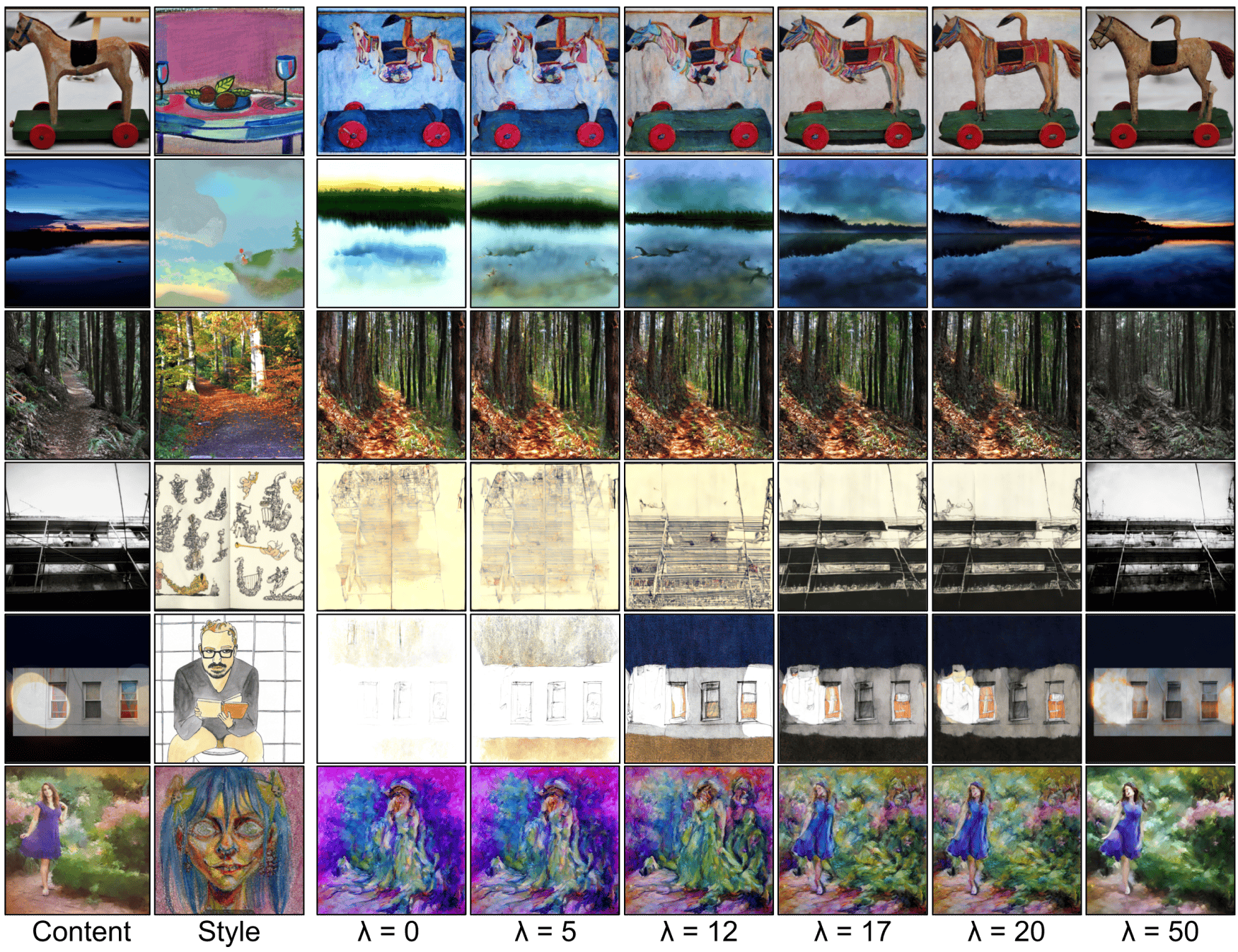}
    \caption{Images generated with different $\lambda$ values. Higher values of lambda lead to more faithfulness in the structure and fine-grained details from the content input. Low lambda values lead to more stylised images that allow more creative structures and flexibility in fine-grained content details. In this example, $T = 50$, meaning lambda can take values from 0 to 50.}
    \label{fig:lambda}
\end{figure*}

 PARASOL offers control in the amount of fine-grained content details that should be kept in the generated image vs. how much the image should be adapted to the new style. This can be controlled via the parameter $\lambda$ (Fig. \ref{fig:lambda}).

 \subsection{Images Generated with Different Classifier-Free Guidance Parameters}

\begin{figure*}[t]
    \centering
    \includegraphics[width=\textwidth]{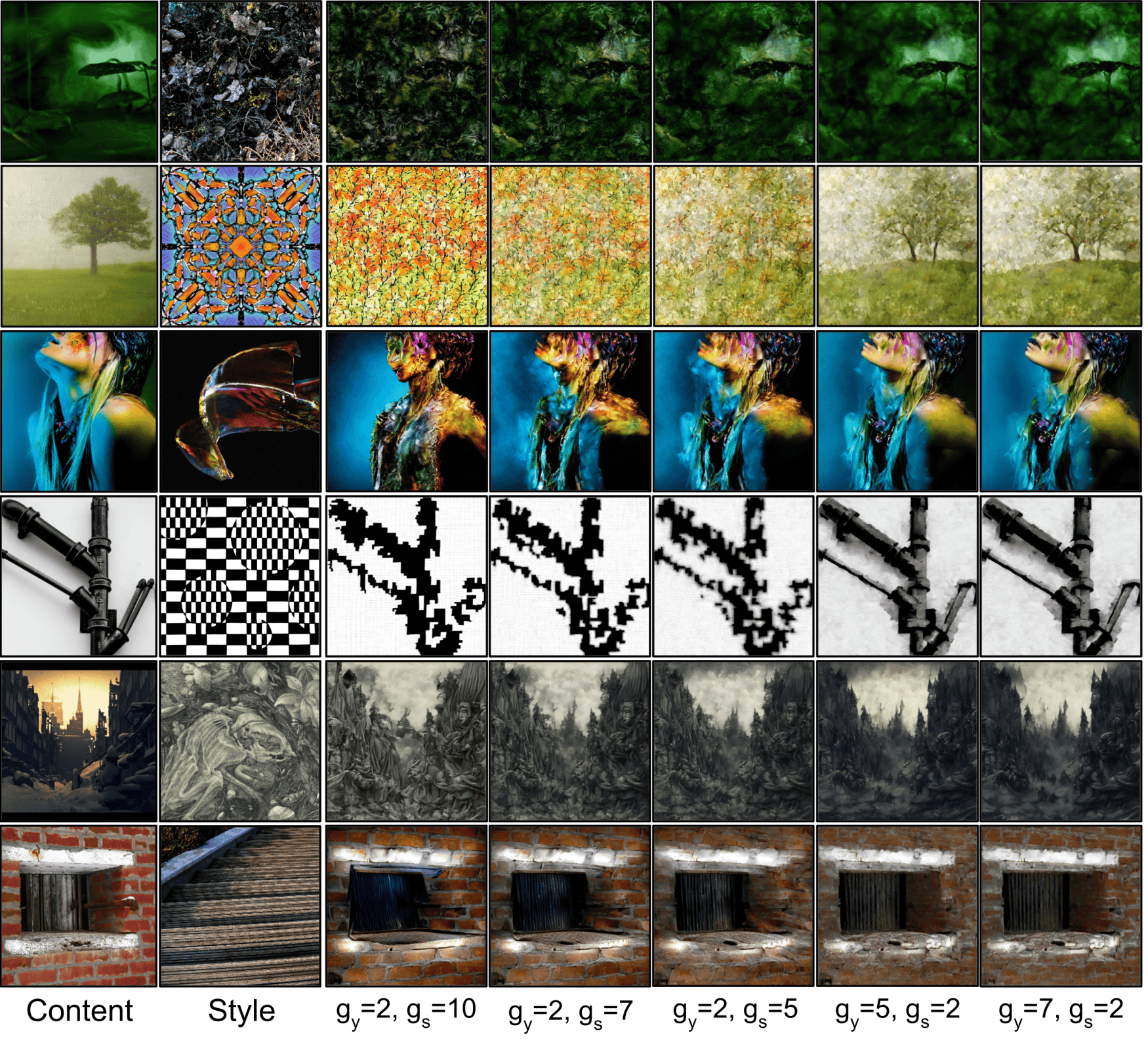}
    \caption{Images generated by PARASOL using different values for $g_s$ and $g_y$. Fixing $\lambda$ means that the structure of the content image is preserved in the same degree for all images. However, the balance between preserved semantics and style change with the ratio of both parameters.}
    \label{fig:gsgy}
\end{figure*}

 The classifier-free guidance parameters $g_s$ and $g_y$ indicate how much weight the style $s$ and content $y$ conditions should have in the generation of the new image. Fig. \ref{fig:gsgy} visualizes the difference those values can make in the generated samples.

  \subsection{Images Generated by combining Different Lambda Values and Classifier-Free Guidance Parameters}

\begin{figure*}[t]
    \centering
    \includegraphics[width=\textwidth]{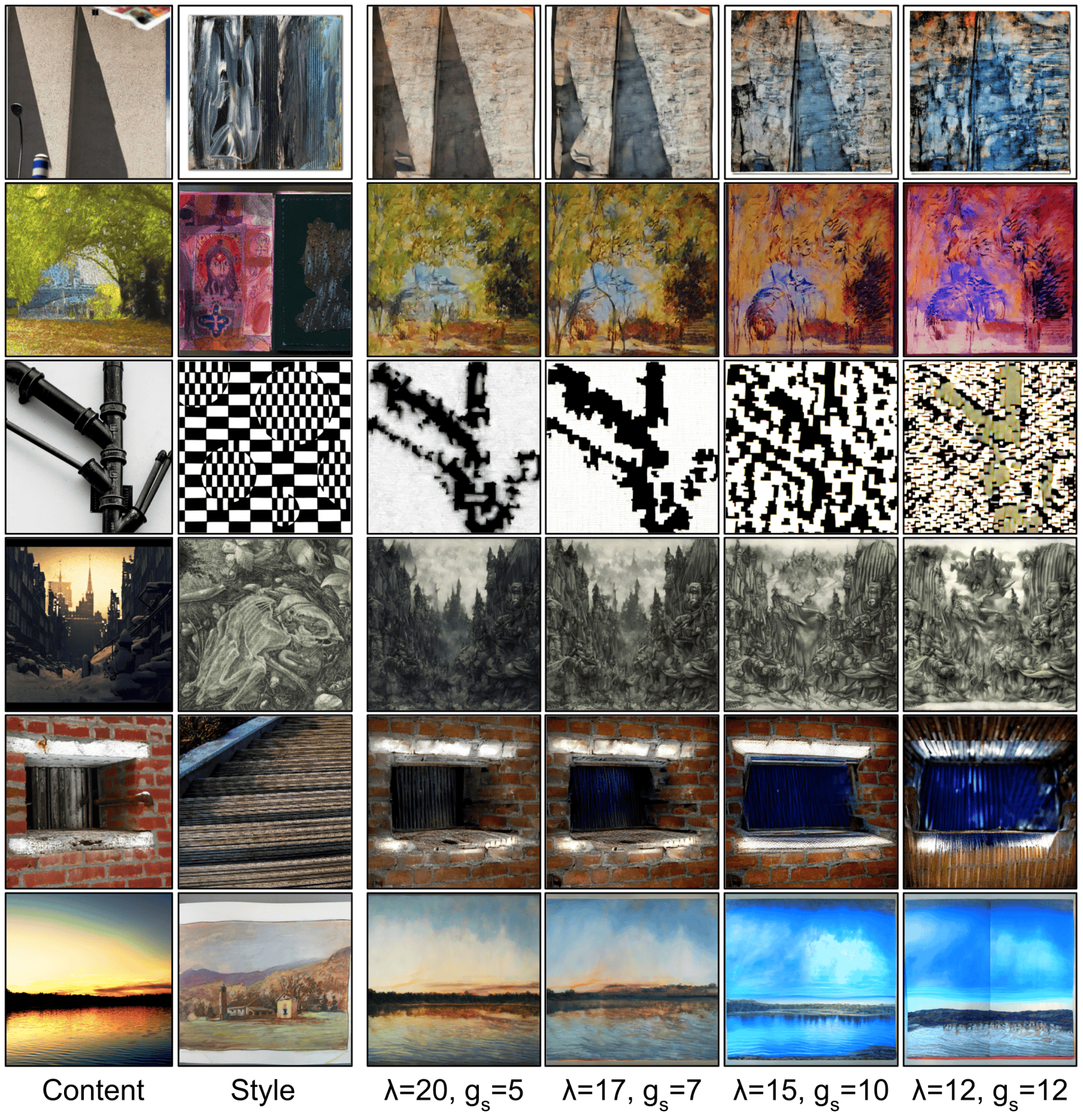}
    \caption{Images generated by PARASOL using different values for $\lambda$ and $g_s$. High values of $\lambda$ paired with low $g_s$ lead to more faithful structures to the content input with more subtle stylization, while high values of $g_s$ and low $\lambda$ values lead to a more noticeable influence of the style image, with more space for creativity in terms of content details. The combination of both parameters allows for a wider range of options in terms of fine-grained controllability of the style and content details in the output image.}
    \label{fig:lambdags}
\end{figure*}

Fig. \ref{fig:lambdags} shows images generated by considering different pairs of $g_s$ and $\lambda$ values, while keeping $g_y$ constant. The ratio of these two parameters defines how much creativity the model is allowed to introduce in the structure and stylization of the image.

\subsection{Images Generated with Style and Content Interpolation}

\begin{figure*}[t]
    \centering
    \includegraphics[width=\textwidth]{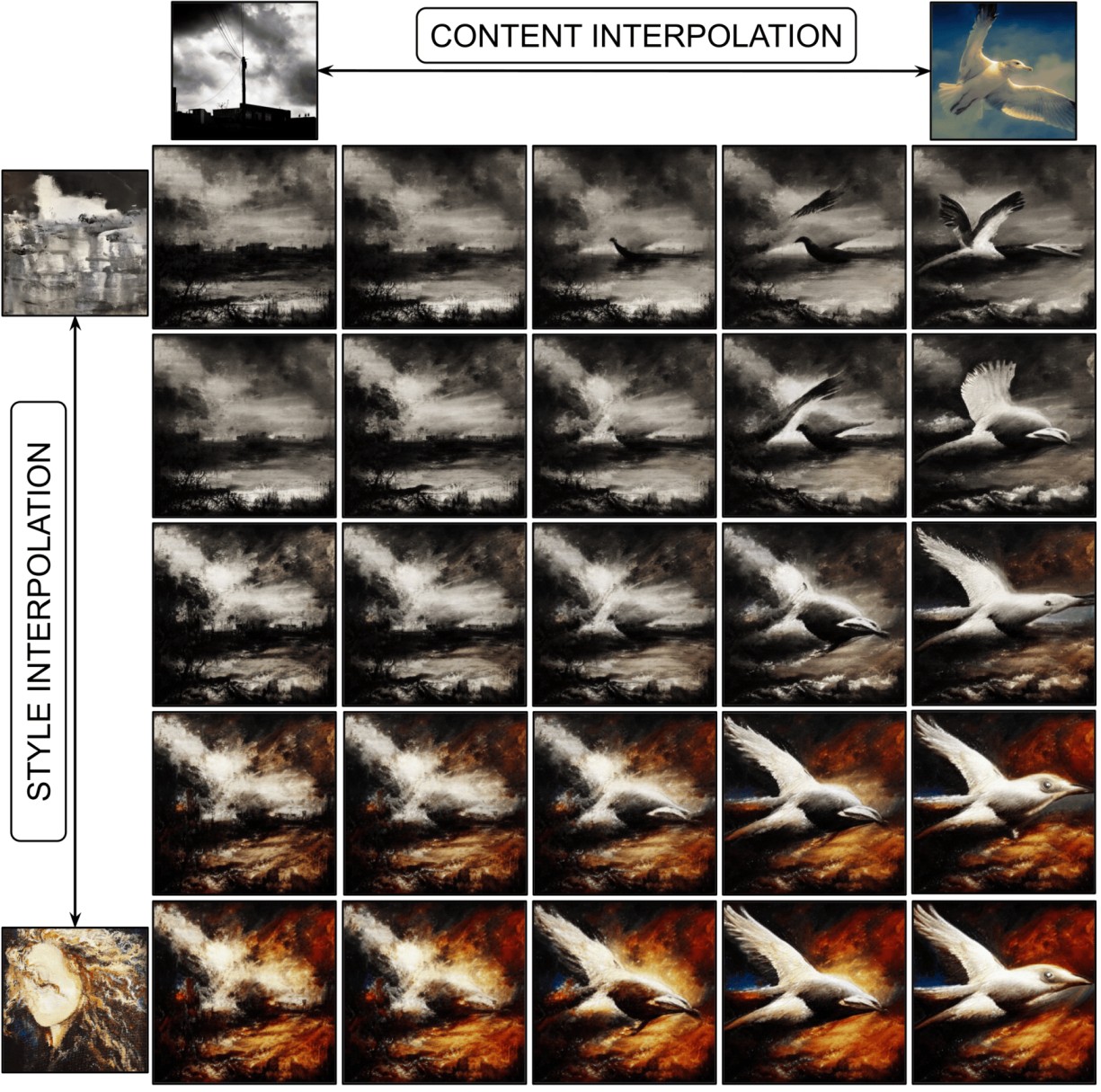}
    \caption{Style and Content interpolations. Visualization of different degrees of interpolation between two content images and two styles. For both style and content interpolations, values $\alpha = 0, 0.25, 0.5, 0.75, 1$ are considered.}
    \label{fig:grid1}
\end{figure*}

\begin{figure*}[t]
    \centering
    \includegraphics[width=\textwidth]{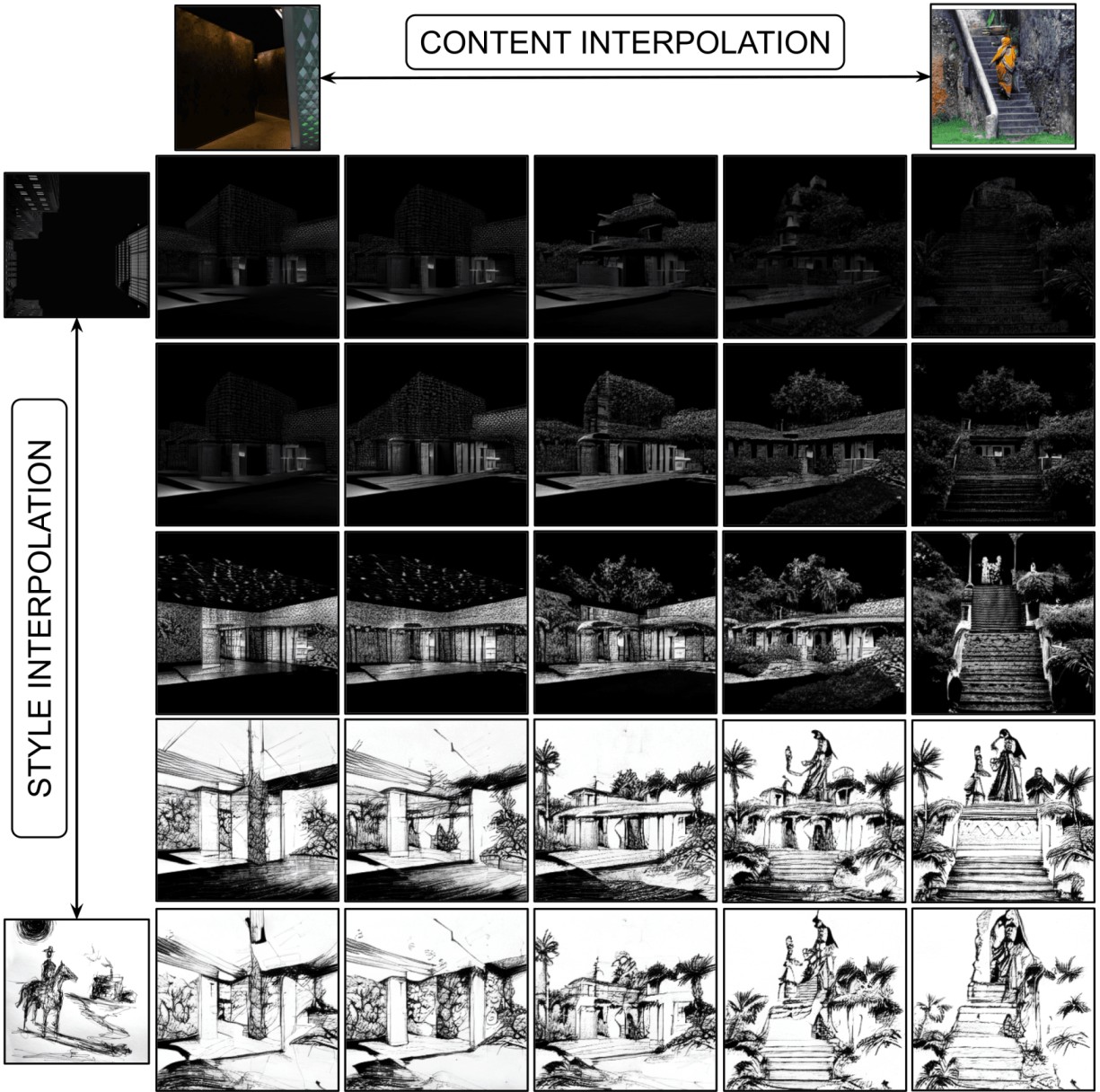}
    \caption{Style and Content interpolations. Second example of images generated by interpolating two content images and two styles in different degrees. For both style and content interpolations, values $\alpha = 0, 0.25, 0.5, 0.75, 1$ are considered.}
    \label{fig:grid2}
\end{figure*}

PARASOL allows the generation of images from a combination of different styles and/or contents (Fig. \ref{fig:grid1}, \ref{fig:grid2}). For combining the information from each pair of descriptors, their spherical interpolation is computed, considering a parameter $0 \le \alpha \le 1$. If $\alpha = 0$ only the first descriptor is considered, while $\alpha = 1$ indicates the second descriptor is the only one taken into account. 

   \subsubsection{Style Interpolation}

\begin{figure*}[t]
    \centering
    \includegraphics[width=\textwidth]{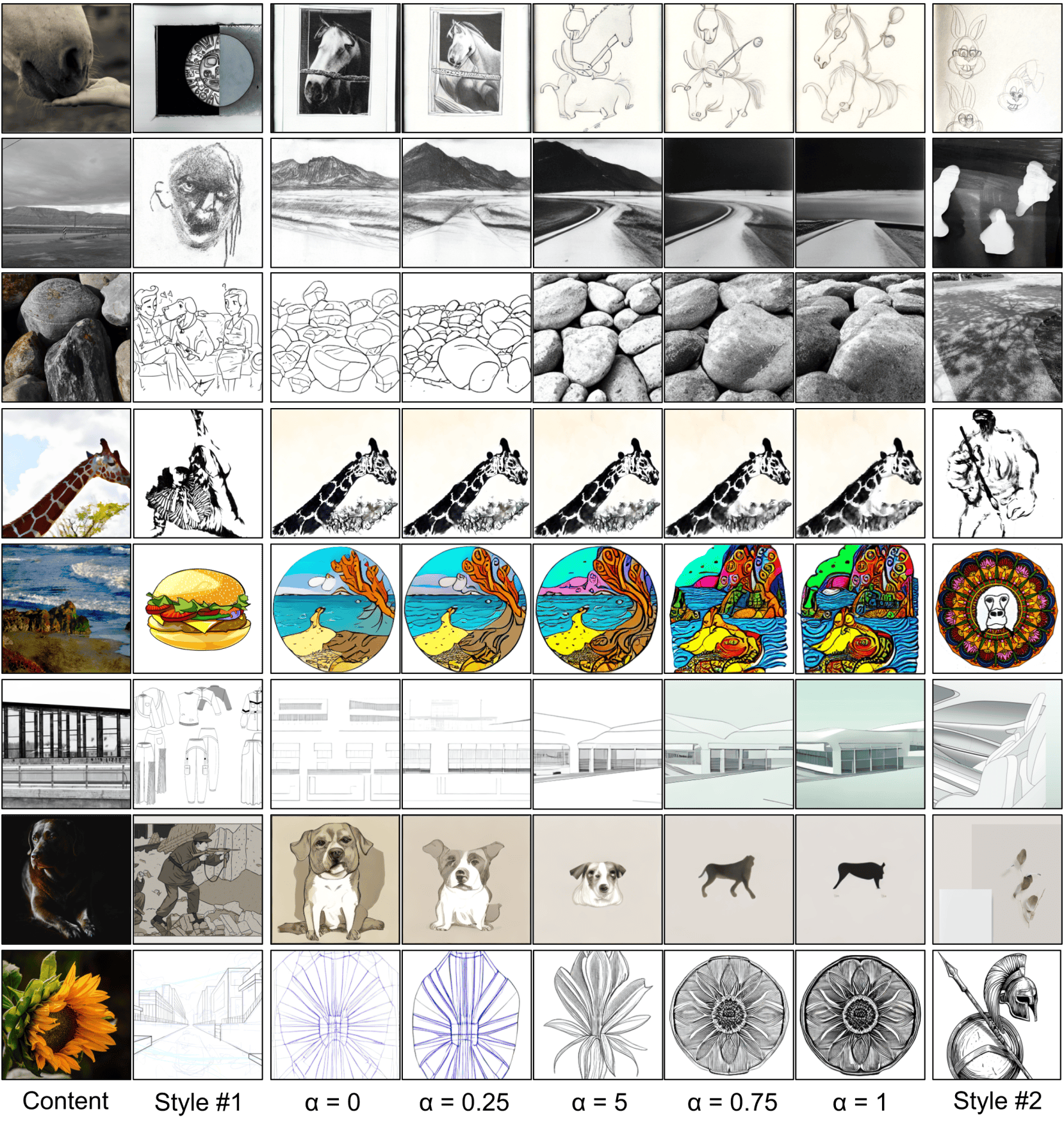}
    \caption{Style interpolation with similar fine-grained styles. Images generated by conditioning on a content image and different interpolations of two very similar styles, transferring and combining their nuances and particular characteristics.}
    \label{fig:interpolationstyle_fg}
\end{figure*}

\begin{figure*}[t]
    \centering
    \includegraphics[width=\textwidth]{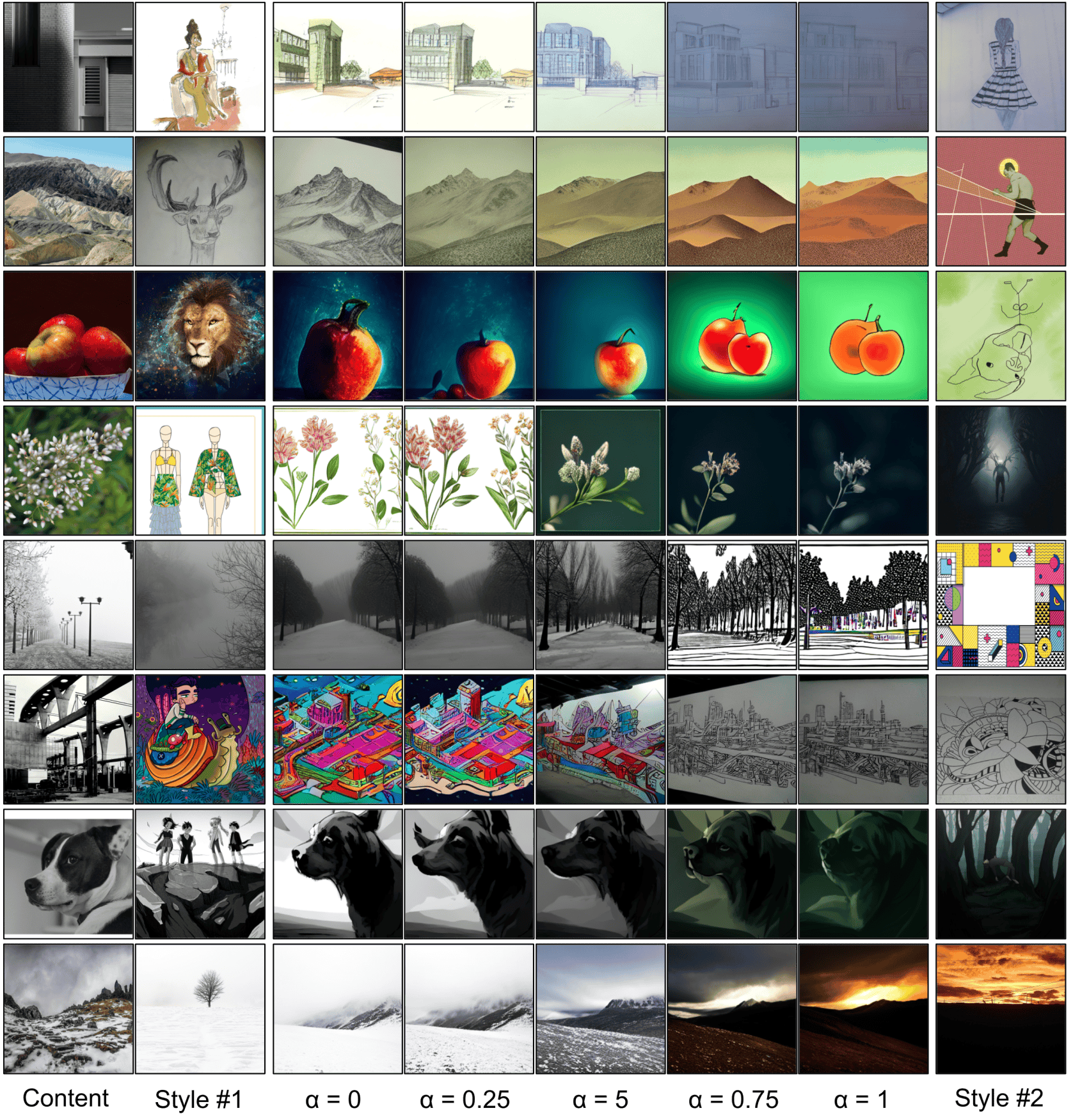}
    \caption{Style interpolation with visually different styles. Images generated by conditioning on a content image and different interpolations of two significantly different styles. PARASOL is able to smoothly transition between both styles, demonstrating its creative capabilities.}
    \label{fig:interpolationstyle_cg}
\end{figure*}

The use of a parametric model \cite{aladin} for encoding the style condition allows the synthesis of images by interpolating different styles. The nuanced information this model is capable to encode allows the interpolation of very similar styles (Fig. \ref{fig:interpolationstyle_fg}) while also being able to interpolate more different styles (Fig. \ref{fig:interpolationstyle_cg}).

\subsubsection{Content Interpolation}

\begin{figure*}[t]
    \centering
    \includegraphics[width=\textwidth]{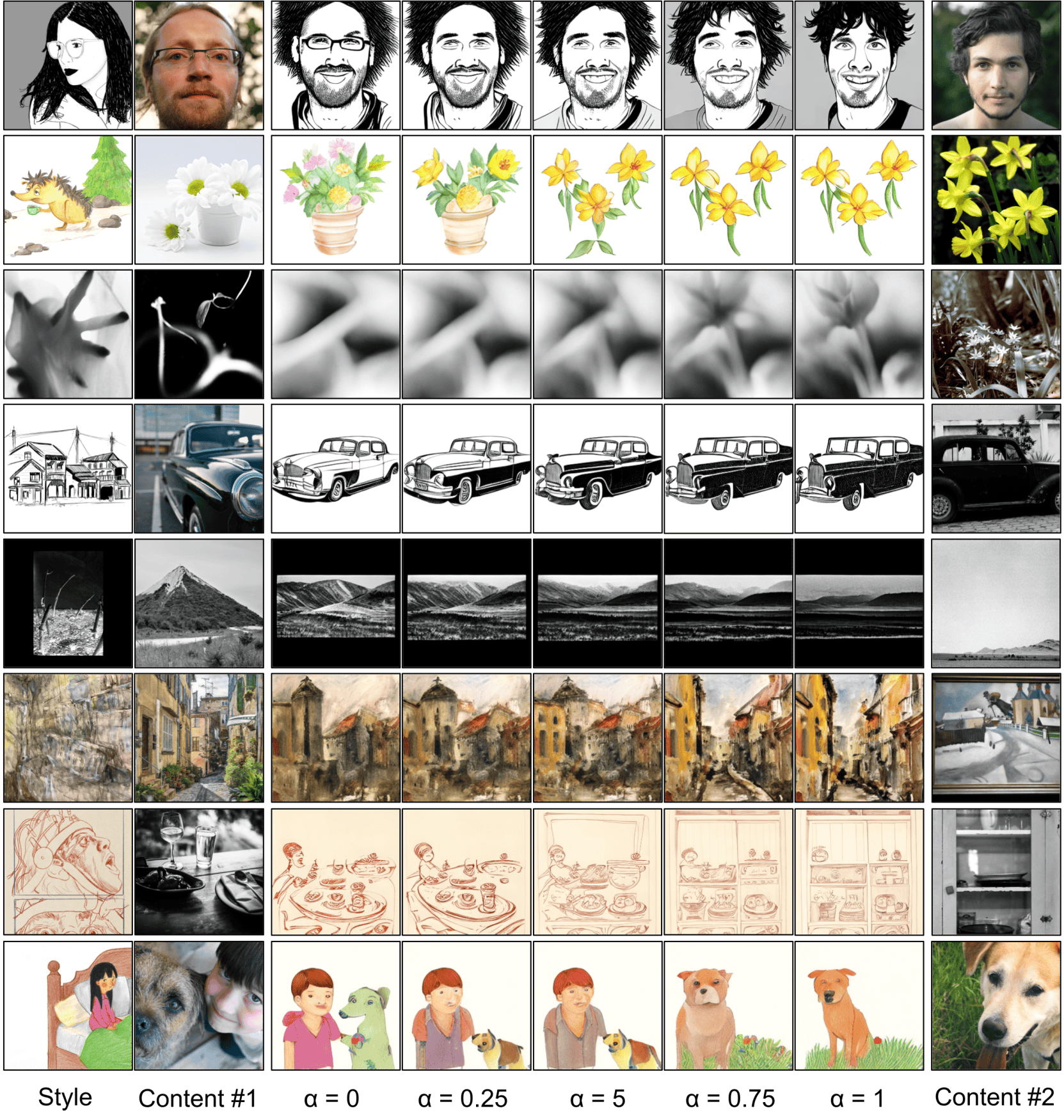}
    \caption{Content interpolation with similar semantics. Images generated by conditioning on a style and content information corresponding to the interpolation of two signals with similar semantics. PARASOL can capture the nuances and fine-grained details of each content input and combine them for generating brand new images.}
    \label{fig:interpolationcontent_fg}
\end{figure*}

\begin{figure*}[t]
    \centering
    \includegraphics[width=\textwidth]{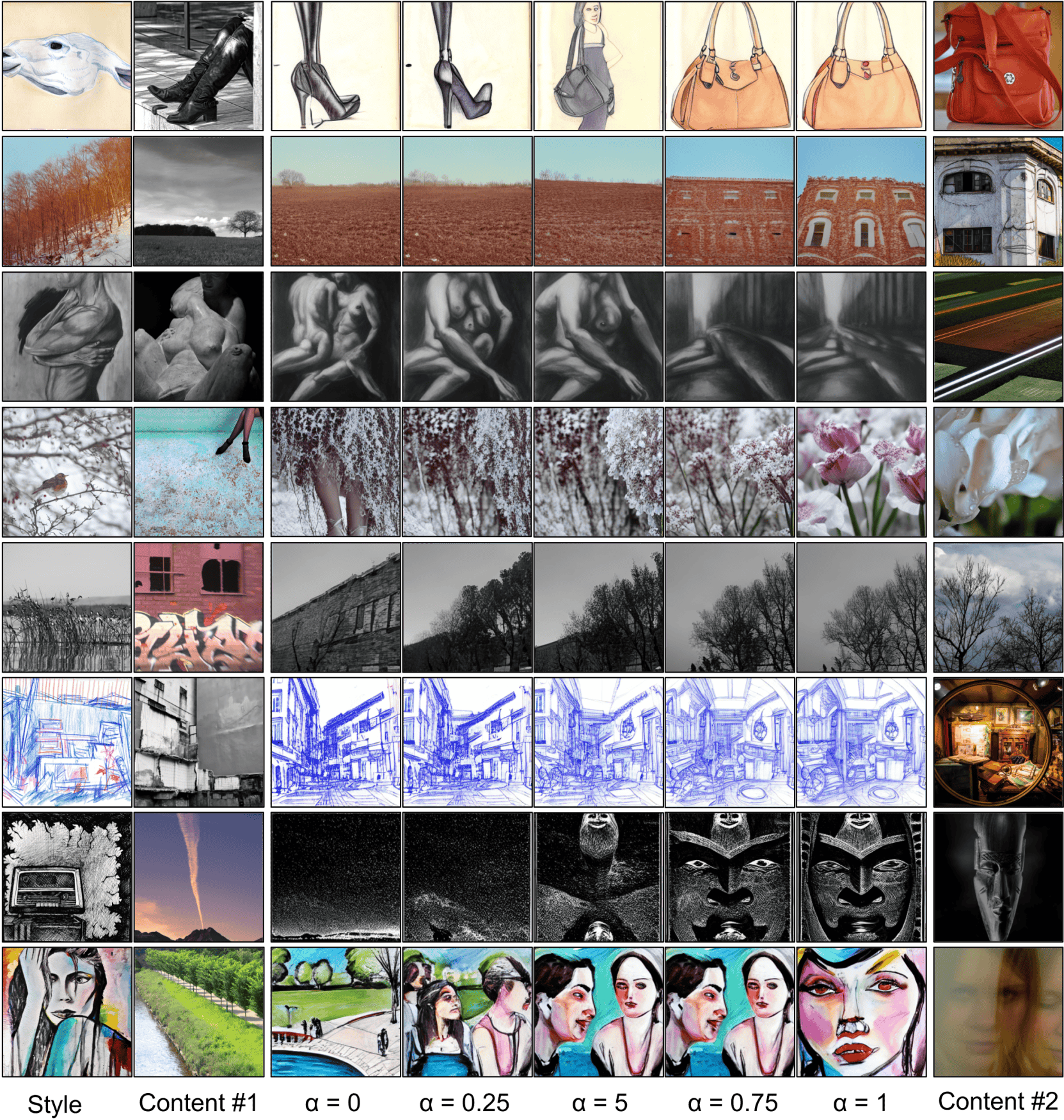}
    \caption{Content interpolation combining different semantics. Images generated by conditioning on a style and content information coming from the interpolation of two signals with different semantics. PARASOL can consistently transfer the style to both scenes or elements and combine them in a smooth way without losing image quality or realism.}
    \label{fig:interpolationcontent_cg}
\end{figure*}

 We encode the content information using CLIP \cite{CLIP} which is also a parametric model. Therefore, not only PARASOL can generate images by interpolating different styles, but it also allows the interpolation of different content information. The content information being interpolated can contain similar semantics (Fig. \ref{fig:interpolationcontent_fg}) or different ones (Fig. \ref{fig:interpolationcontent_cg}).

 \subsection{Images Generated with Different Fine-Grained Content Details}

\begin{figure*}[t]
    \centering
    \includegraphics[width=\textwidth]{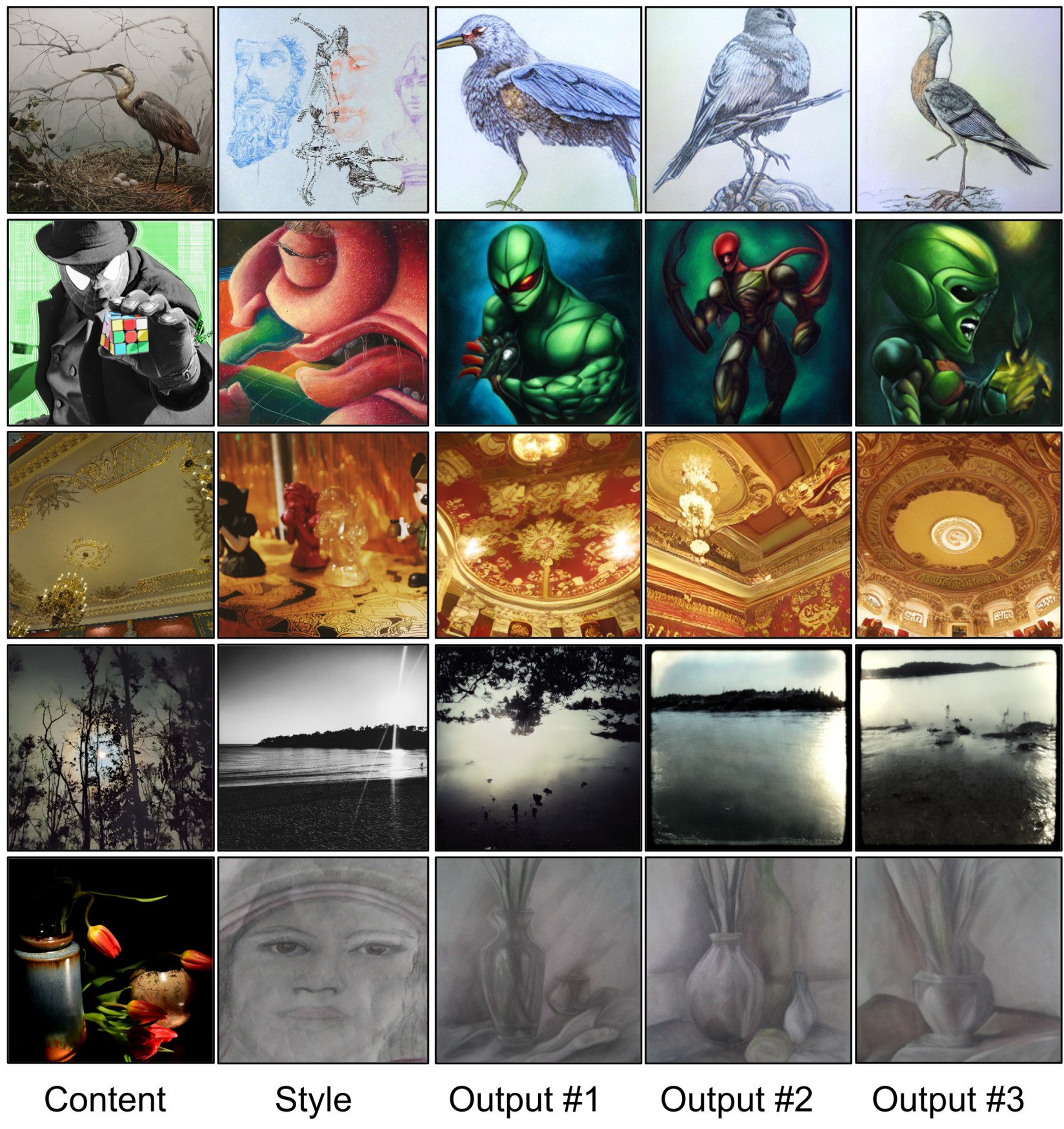}
    \caption{Diversity in fine-grained content. Images generated by PARASOL allowing flexibility in the fine-grained content details and image structure.}
    \label{fig:diversity}
\end{figure*}

Section \ref{seq:controllability} (C) details how PARASOL can generate images with consistent semantics and fine-grained style while offering diversity in the fine-grained content details. Fig. \ref{fig:diversity} offers examples of this use case using $\lambda=20$, $g_s=5$ and $g_y=5$. However, those parameters could be tuned for further control over the attributes of the final image.

